\documentclass[sigconf]{acmart}

\usepackage{booktabs} 
\usepackage{graphicx}
\usepackage{subcaption}
\usepackage{algorithm}
\usepackage{algorithmic}
\usepackage{harpoon}
\usepackage{multicol}
\usepackage{multirow}
\usepackage{amsmath}
\usepackage{mathtools}
\usepackage{amsfonts}
\usepackage{color}
\usepackage{enumitem}
\usepackage{footmisc}
\usepackage{verbatim}

\def\BibTeX{{\rm B\kern-.05em{\sc i\kern-.025em b}\kern-.08emT\kern-.1667em\lower.7ex\hbox{E}\kern-.125emX}}
    
\copyrightyear{2019}
\acmYear{2019}
\setcopyright{none} 

\acmArticle{4}

\begin{document}

%
\title[Parallel Locality-Optimized NMF]{PL-NMF: Parallel Locality-Optimized Non-negative Matrix Factorization}

%
\author{Gordon E. Moon}
\affiliation{
  \institution{The Ohio State University}
  \city{Columbus}
  \state{Ohio}
  \postcode{43210}
  \country{U.S.A}
}
\email{moon.310@osu.edu}
\author{Aravind Sukumaran-Rajam}
\affiliation{
  \institution{The Ohio State University}
  \city{Columbus}
  \state{Ohio}
  \postcode{43210}
  \country{U.S.A}
}
\email{sukumaranrajam.1@osu.edu}
\author{Srinivasan Parthasarathy}
\affiliation{
  \institution{The Ohio State University}
  \city{Columbus}
  \state{Ohio}
  \postcode{43210}
  \country{U.S.A}
}
\email{srini@cse.ohio-state.edu}
\author{P. Sadayappan}
\affiliation{
  \institution{The Ohio State University}
  \city{Columbus}
  \state{Ohio}
  \postcode{43210}
  \country{U.S.A}
}
\email{sadayappan.1@osu.edu}

%
\renewcommand{\shortauthors}{G. Moon et al.}

%
\begin{abstract}
Non-negative Matrix Factorization (NMF) is a key kernel for
unsupervised dimension reduction used in a wide range of applications,
including topic modeling, recommender systems and bioinformatics.
Due to the compute-intensive nature of applications that must perform repeated
NMF, several parallel implementations have been developed in the past. However,
existing parallel NMF algorithms have not addressed data locality optimizations,
which are critical for high performance since data movement costs greatly exceed
the cost of arithmetic/logic operations on current computer systems.
In this paper, we devise a parallel NMF algorithm based on the HALS
(Hierarchical Alternating Least Squares) scheme that incorporates algorithmic
transformations to enhance data locality. Efficient realizations of the algorithm
on multi-core CPUs and GPUs are developed, demonstrating significant performance
improvement over existing state-of-the-art parallel NMF algorithms.
\end{abstract}

%
%


%

%

%
\maketitle

\section{Introduction}
%

Non-negative Matrix Factorization (NMF) is a key primitive used in a
wide range of applications, including topic modeling \cite{shi2018short,kuang2015nonnegative,suh2017local}, recommender systems \cite{hernando2016non,aghdam2015novel,zhang2006learning} and bioinformatics \cite{yang2015non,wang2013non,mejia2015nmf}. Given a non-negative matrix $A \in \mathbb
R_+^{V \times D}$ and $K \ll \textrm{ min}(V,D)$, NMF finds two
non-negative rank-K matrices $W \in \mathbb R_+^{V \times K}$ and
$H \in \mathbb R_+^{K \times D}$, such that the product of $W$ and $H$
approximates $A$ \cite{lee2001algorithms}:
\begin{equation}
\label{eq:NMF}
A \approx W H
\end{equation}
NMF is a powerful technique for topic modeling. When $A$ is a corpus in which
each document is represented as a collection of bag-of-words from an active
vocabulary, the factor matrices $W$ and $H$ can be interpreted as
latent topic distributions for words and documents.

Several algorithms have been proposed for NMF. They all involve repeated
alternating update of some elements of $W$ interleaved with update of some
elements of $H$, with imposition on non-negativity constraints on the elements,
until a suitable error norm (either Frobenius norm or Kullback-Leibler
divergence) is lower than a desired threshold. Various previously developed
algorithms for NMF differ in the granularity of the number of elements of $W$
that are updated before switching to updating some elements of $H$. The focus of
prior work has been to compare the rates of convergence of alternate algorithms
and the parallelization of the algorithms.
However, to the best of our knowledge, the minimization of data movement through
the memory hierarchy, using techniques like tiling, has not been previously
addressed.  With costs of data movement from memory being significantly higher
than the cost of performing arithmetic operations on current processors, data
locality optimization is extremely important. 

In this paper, we address the issue of data locality optimization for NMF.
An analysis of the computational components
of the FAST-HALS (Hierarchical Alternating Least Squares) algorithm
for NMF \cite{cichocki2009fast}, 
is first performed to identify data movement overheads.
The associativity of addition is
utilized to judiciously reorder additive contributions in updating elements of
$W$ and $H$, to enable 3D tiling of a computationally intensive component of
the algorithm. 
An analysis of the data movement
overheads as a function of tile size is developed, leading to a model for
selection of effective tile sizes. Parallel implementations of the new \textbf{P}arallel \textbf{L}ocality-optimized \textbf{NMF} algorithm (called \textbf{PL-NMF}) are presented for both GPUs and multi-core CPUs. An
experimental evaluation with datasets used in prior studies demonstrates
significant performance improvement over state-of-the-art alternatives available
for parallel NMF.

The paper is organized as follows. In the next section, we present the
background on NMF and related prior work. In Section 3, we present the
high-level overview of PL-NMF algorithm. Sections 4 and 5 demonstrate details of
our PL-NMF for multi-core CPUs and GPUs.  In Section 6, we systematically analyze
the data movement cost for PL-NMF and original FAST-HALS algorithms. Section 7
presents determination of the tile sizes based on data movement analysis. In
Section 8, we compare PL-NMF with existing state-of-the-art parallel
implementations.

\section{Background and Related Work}
\label{section:background_related_work}
\subsection{Non-negative Matrix Factorization Algorithms}

NMF seeks to solve the optimization problem of minimizing reconstruction
error between $A$ and the approximation $WH$. In order to measure the reconstruction error for NMF, Lee et al. \cite{lee2001algorithms} adopted
various objective functions, such as the Frobenius norm given two
matrices and Kullback-Leibler divergence given two probability distributions.
The objective functions $D(A||W H)$ based on the Frobenius norm is defined in Equation \ref{eq:frobenius_norm}. 
\begin{equation} \label{eq:frobenius_norm} \text{$D_{F}(A||W H) =
	\frac{1}{2}||A - W H||_{F}^{2} =  \frac{1}{2}\sum_{vd} \left(A_{vd}-(W
H)_{vd}\right)^{2}$} 
\end{equation}
To efficiently minimize the objective functions (above), several variants of NMF
algorithms have been developed: 
{\em Multiplicative Update (MU)}, {\em Additive Update (AU)}, {\em Alternating
Non-negative Least Squares (ANLS)} and {\em Hierarchical Alternating Least
Squares (HALS)}. Table \ref{tb:nmf_notations} describes the notations used in this
paper.
\begin{table}[h]
\centering
\caption{Common notations for NMF algorithms}
\label{tb:nmf_notations}
\scalebox{0.8}{
\begin{tabular}{|c|l|}
\hline
\textbf{Notation} & \multicolumn{1}{c|}{\textbf{Description}}       \\ \hline
$A$        & Non-negative matrix         \\ \hline
$W$        & Non-negative rank-$K$ matrix factor         \\ \hline
$H$        & Non-negative rank-$K$ matrix factor    \\ \hline
$V$        & Number of rows in $A$ and $W$ \\ \hline
$D$        & Number of columns in $A$ and $H$     \\ \hline
$K$        & Low rank                       \\ \hline
\end{tabular}
}
\end{table}

Multiplicative update (MU) and additive update (AU) proposed by Lee et al.
\cite{lee2001algorithms} are the simplest NMF algorithms. The MU algorithm
updates two rank-$K$ non-negative matrices $W$ and $H$ based on multiplicative
rules and ensures convergence. MU strictly conforms to non-negativity
constraints on $W$ and $H$ because the elements of $W$ and $H$ that have zero
value will not be updated. Unlike MU algorithm, the AU algorithm updates $W$ and
$H$ based on the gradient descent method and avoids negative update values using learning rate. However, some studies have
reported that the use of MU and AU algorithms leads to weaknesses such as slower
convergence and lower convergence rate
\cite{gonzalez2005accelerating,lin2007projected,kim2008nonnegative}.

Alternating Non-negative Least Squares (ANLS) is a special type of Alternating
Least Squares (ALS) approach. At each iteration, the gradients of two objective
functions with respect to $W$ and $H$ are used to update each of $W$ and $H$ one
after the other.  Kim et al. \cite{kim2011fast} proposed Alternating
Non-negative Least Squares based Block Principle Pivoting (ANLS-BPP) algorithm.
Under the Karush-Kuhn-Tucker (KTT) conditions, the ANLS-BPP algorithm
iteratively finds the indices of non-zero elements (passive set) and zero
elements (active set) in the optimal matrices until KTT conditions are
satisfied. The values of indices that correspond to the active set will become
zero, and the values of passive set are approximated by solving $min||A - W
H||_{F}^{2}$ which is a standard Least Squares problem.

As an alternative to the basic ANLS approach, Cichocki et al.
\cite{cichocki2007hierarchical} proposed Hierarchical Alternating Least Squares
(HALS), which hierarchically updates only one $k$-th row vector of $H \in \mathbb
R_+^{K \times D}$ at a time and then uses it to update a corresponding $k$-th
column vector of $W \in \mathbb R_+^{V \times K}$. In other words, HALS
minimizes the $K$ set of two local objective functions with respect to $K$ row
vectors of $H$ and $K$ column vectors of $W$ at each iteration. A standard HALS
algorithm iteratively updates each row of $H$ and each column of $W$ in order
within the innermost loop.

Based on the standard HALS algorithm, Cichocki et al.  \cite{cichocki2009fast}
further proposed the extended version of a new algorithm called FAST-HALS
algorithm as described in Algorithm \ref{alg:FAST_HALS}. Note that $H_{k}$ and
$W_{k}$ indicate $k$-th row of $H$ and $k$-th column of $W$, respectively.
FAST-HALS updates all rows of $H$ before starting the update to all columns of
$W$, instead of alternately updating each row of $H$ and each column of $W$ at a
time. Compared to MU algorithm, the FAST-HALS algorithm converges much faster
and produces a better solution, while maintaining a similar computational cost
as reported in \cite{kim2011fast,gillis2014and}. Interestingly, Kim et al.
\cite{kim2011fast} have shown that FAST-HALS has also been found to converge
faster than their ANLS-BPP implementation on real-world text datasets: TDT2 and
20 Newsgroups, while maintaining the same convergence rate (see Figure 5.3 in
Kim et al. \cite{kim2011fast}).

\subsection{Related Work on Parallel NMF}
Since most of the variations of NMF algorithm are highly compute-intensive, many
previous efforts have been made to parallelize NMF algorithms. As shown in Table
\ref{tb:prev_nmf}, previous studies on parallelizing NMF can be broadly
categorized into two groups based on implementation for multi-core CPUs
\cite{battenberg2009accelerating,fairbanks2015behavioral,dong2010parallel,liu2010distributed,liao2014cloudnmf,kannan2016high}
versus GPUs \cite{lopes2010non,mejia2015nmf,koitka2016nmfgpu4r}. Furthermore,
each study used various NMF algorithms for parallel implementations.

\begin{table}[h]
\centering
\caption{Previous studies on parallelization of NMF}
\label{tb:prev_nmf}
\scalebox{0.8}{
\begin{tabular}{|c|c|c|c|}
\hline
\textbf{Author}                   & \textbf{Machine}          &\textbf{Platform}			  & \textbf{Algorithm} \\ \hline
Battenberg et al. \cite{battenberg2009accelerating} & CPU                       & Shared-memory                 & MU        \\ \hline
Fairbanks et al. \cite{fairbanks2015behavioral}    & CPU                       & Shared-memory                 & ANLS-BPP  \\ \hline
Dong et al. \cite{dong2010parallel}           & CPU                       & Distributed-memory            & MU        \\ \hline
Liu et al. \cite{liu2010distributed}         & CPU                       & Distributed-memory            & MU        \\ \hline
Liao et al. \cite{liao2014cloudnmf}           & CPU                       & Distributed-memory            & MU        \\ \hline
Kannan et al. \cite{kannan2016high}             & CPU                       & Distributed-memory            & ANLS-BPP  \\ \hline
Lopes et al. \cite{lopes2010non}               & GPU                       & Shared-memory                 & MU, AU  \\ \hline
Koitka et al. \cite{koitka2016nmfgpu4r}         & GPU                       & Shared-memory                 & MU, ALS  \\ \hline
Mej{\'\i}a-Roa et al. \cite{mejia2015nmf}               & GPU                       & Distributed-memory            & MU  \\ \hline
\end{tabular}
}
\end{table}

\subsubsection{Shared-Memory Multiprocessor}
\hfill\\
Battenberg et al. \cite{battenberg2009accelerating} introduced parallel NMF
using MU algorithm for audio source separation task. Fairbanks et al.
\cite{fairbanks2015behavioral} adopted ANLS-BPP based NMF in order to find the
structure of temporal behavior in a dynamic graph given vertex features. Both
\cite{battenberg2009accelerating} and \cite{fairbanks2015behavioral} developed the
parallel NMF implementations on multi-core CPUs using Intel Math Kernel Library
(MKL) along with shared-memory multiprocessor.
\subsubsection{Distributed-Memory Systems} \hfill\\
Dong et al. \cite{dong2010parallel} demonstrated that MU algorithm and
shared-memory based parallel implementation have a limitation of slow
convergence. To overcome these problems, they devised a parallel MPI
implementation of MU based NMF that improves Parallel NMF (PNMF) proposed by
Robila et al. \cite{robila2006parallel}. Different NMF algorithms have
previously used tiling/blocking to minimize data movement. Dong et al.
\cite{dong2010parallel} partitioned the two factor matrices, W and H, into
smaller blocks and each block is distributed to different threads. Each block
simultaneously updates corresponding sub-matrices of the two matrices, and a
reduction operation is performed by collective communication operations using
Message Passing Interface (MPI). Similarly, Liu et al. \cite{liu2010distributed}
proposed matrix partition scheme that partitions the two factor matrices along
the shorter dimension ($K$ dimension) instead of the longer dimensions ($V$ or
$D$ dimensions). Therefore, each matrix is divided up to more partitions
compared to partitioning along the longer dimension, so that the data locality
is increased and the communication cost is decreased when performing the product
of two matrices. Kannan et al. \cite{kannan2016high} minimized the communication
cost by communicating only with the two factor matrices and other partitioned
matrices among parallel threads. Based on the ANLS-BPP algorithm, their
implementation also reduced the bandwidth and data latency using MPI collective
communication operations. Given an input matrix $A$ and two factorized matrices
$W$ and $H$, they partitioned $W$ and $H$ into $P$ multiple blocks (tiles)
across $V$ and $D$ dimensions which are the number of rows in $W$ and columns in
$H$. Hence, the sizes of each block in $W$ and $H$ are ($V$/$P$)$\times$ $K$ and
$K$ $\times$($D$/$P$), respectively. Doing so allows the matrix $A$ to be
partitioned into $P$ tiles $\times$ $P$ tiles. Then $P$ different processors
perform matrix multiplication with the different $P$ tiles of $W$ and $H$
simultaneously. This data partition scheme is appropriate for block-wise updates
of $W$ and $H$ based on ANLS-BPP algorithm. Unlike ANLS-BPP algorithm, FAST-HALS
requires column-wise/row-wise sequential updates because there is a data
dependency between two consecutive columns/rows. Hence, FAST-HALS algorithm is
not allowed to divide $W$ and $H$ across $V$ and $D$ dimensions. In our tiling
approach, $W$ and $H$ are partitioned across $K$ dimension, and the sizes of
each block in $W$ and $H$ are $V$ $\times$($K$/$P$) and ($K$/$P$)$\times$ $D$,
respectively. Our key contribution is not tiling/blocking itself, but converting
matrix-vector operations to matrix-matrix operations. Tiling enables us to do
the latter.

\subsubsection{GPU Platform}
\hfill\\
Lopes et al. \cite{lopes2010non} proposes several GPU-based parallel NMF
implementations that use both MU and AU algorithms for both Euclidean and KL
divergence objective functions.  Mej{\'\i}a-Roa et al. \cite{mejia2015nmf}
presents NMF-mGPU that performs MU based NMF algorithm on either a single GPU
device or multiple GPU devices through MPI for a large-scale biological dataset.
Koitka et al. \cite{koitka2016nmfgpu4r} presents MU and ALS based GPU
implementations binding to the R environment.  To our knowledge, our paper is
the first to develop FAST-HALS based parallel NMF implementation for GPUs.

\section{Overview of Approach}
\label{section:overview_scheme}

In this section, we present a high-level overview of our approach to optimize
NMF for data locality. We begin by describing the FAST-HALS algorithm 
\cite{cichocki2009fast}, one of the fastest algorithms for NMF
as demonstrated by previous comparison studies \cite{kim2011fast}.
We analyze the data movement overheads from main memory, for different components of
that algorithm, and identify the main bottlenecks. We then show how
the algorithm can be adapted by exploiting the associativity of
addition to make the computation effectively tileable to reduce
data movement from memory, whereas
the original form is not tileable.

\subsection{Overview of FAST-HALS Algorithm}
\begin{algorithm}
	\caption{FAST-HALS algorithm}
	\label{alg:FAST_HALS}
	\begin{flushleft}
		\textbf{Input}: $A \in \mathbb R_+^{V \times D}$: non-negative matrix, $\epsilon$: small non-negative quantity\\
		\end{flushleft}
		\begin{algorithmic}[1]
			\STATE Initialize $W \in \mathbb R_+^{V \times K}$ and $H \in \mathbb R_+^{K \times D}$ with random non-negative numbers \\
			\REPEAT
			\STATE \textcolor{black}{\textbf{// Updating H}}
			\STATE $R$ $\leftarrow$ $A^{T}W$
			\STATE $S$ $\leftarrow$ $W^{T}W$
			\FOR{k $=$ 0 to K $-$ 1}
				\STATE $H_{k}$ $\leftarrow$ max$\left(\epsilon,H_{k} + R_{k} - H^{T}S_{k}\right)$
			\ENDFOR
			\STATE \textcolor{black}{\textbf{// Updating W}}
			\STATE $P$ $\leftarrow$ $AH^{T}$
			\STATE $Q$ $\leftarrow$ $HH^{T}$
			\FOR{k $=$ 0 to K $-$ 1}
				\STATE $W_{k}$ $\leftarrow$ max$\left(\epsilon,W_{k}Q_{kk} + P_{k} - WQ_{k}\right)$
				\STATE // Normalize $W_{k}$ column vector with $L_{2}-norm$
				\STATE $W_{k} \leftarrow \dfrac{W_{k}}{||W_{k}||_{2}}$
			\ENDFOR
			\UNTIL{convergence}
		\end{algorithmic}
\end{algorithm}



Algorithm~\ref{alg:FAST_HALS} shows pseudo-code for the FAST-HALS algorithm
\cite{cichocki2009fast} for NMF. It is an iterative algorithm that
iteratively updates $H$ and $W$, fully updating all entries in $H$ (lines 4-8) and then updating all entries in $W$ (lines 10-15) during each iteration until convergence. While the updates to $H$ and $W$ are slightly different (due to normalization of $W$ after each iteration), each of the updates involves a pair of matrix-matrix products (lines 4/5 and 10/11 for $H$ and $W$, respectively) and a sequential loop that steps through features ($k$ loop) to update one row (column) of $H$($W$) at a time. The computation within these $k$ loops involves vector-vector
operations and matrix-vector operations. From a computational
complexity standpoint, the various matrix-matrix products and the sequential
($K$ times) matrix-vector products all have cubic complexity (O($N^3)$ if all
matrices are square and of side $N$). But as we
show by analysis of data movement requirements in the next sub-section, the
collection of matrix-vector products in lines 7 and 13 dominate. In the
following sub-section, we present our approach to alleviating this bottleneck
by exploiting the flexibility of instruction reordering via use of the
associativity property of addition\footnote{\scriptsize{Floating-point addition is of
course not strictly associative, but as shown later by the experimental results, the changed order does not adversely affect algorithm convergence.}}.

\subsection{Data Movement Analysis for FAST-HALS Algorithm}
The code regions with high data movement can
be identified by individually analyzing each line in Algorithm
\ref{alg:FAST_HALS}. Lines 4 and 5 perform matrix multiplication. 
It is well known that $\frac{2MNK}{\sqrt{C}}$ is the highest order term in the number of data elements moved (between main memory and a cache of size $C$ words) for efficient tiled matrix multiplication of two matrices $A$, ($M \times K$) and $B$, ($K \times N$)\footnote{An extensive discussion of both
lower bounds and data movement volume for several tiling
schemes may be found in the recent work of Smith \cite{smith2018theory}.}. 
Thus, the data movement costs associated with lines 4 and 5 are
$\frac{2DKV}{\sqrt{C}}$ and $\frac{2KKV}{\sqrt{C}}$, respectively.
The loop in line 6 performs matrix-vector multiplication and has an
associated data movement cost of $K(3D+DK+K)$. Similar to lines 4 and 5, the
data movement costs for lines 10 and 11 are 
$\frac{2VKD}{\sqrt{C}}$ and $\frac{2KKD}{\sqrt{C}}$, respectively.
The loop in line 12 has an associated data movement cost of $K(VK + K + 6V +
1)$. The total data movement for Algorithm \ref{alg:FAST_HALS} is shown in
Equation \ref{eq:mat_vec_original_data_movment}.

\begin{equation}
\label{eq:mat_vec_original_data_movment}
K(K(V+D)(1+\frac{2}{\sqrt{C}})+\frac{4VD}{\sqrt{C}}+6V+3D+2K+1)
\end{equation}

The main data movement overhead is associated with loops in lines 6 and 12. For
example, the combined fractional data movement overhead of lines 7 (within loop
in line 6) and 13 (within loop in line 12) is 91\% for the 20 Newsgroups
dataset. If the operational intensity (defined as the number of operations per
data element moved) is very low, the performance will be bounded by memory
bandwidth and thus will not be able to achieve the peak compute capacity. Due to
its low operational intensity, the performance of Algorithm \ref{alg:FAST_HALS}
is limited by the memory bandwidth. Thus, the major motivation for our algorithm
adaptation is to achieve better performance by reducing the required data movement.

\subsection{Overview of PL-NMF}

\begin{figure}
	\centering
	\includegraphics*[viewport= 20 250 550 530, clip=true, width=0.99\columnwidth]{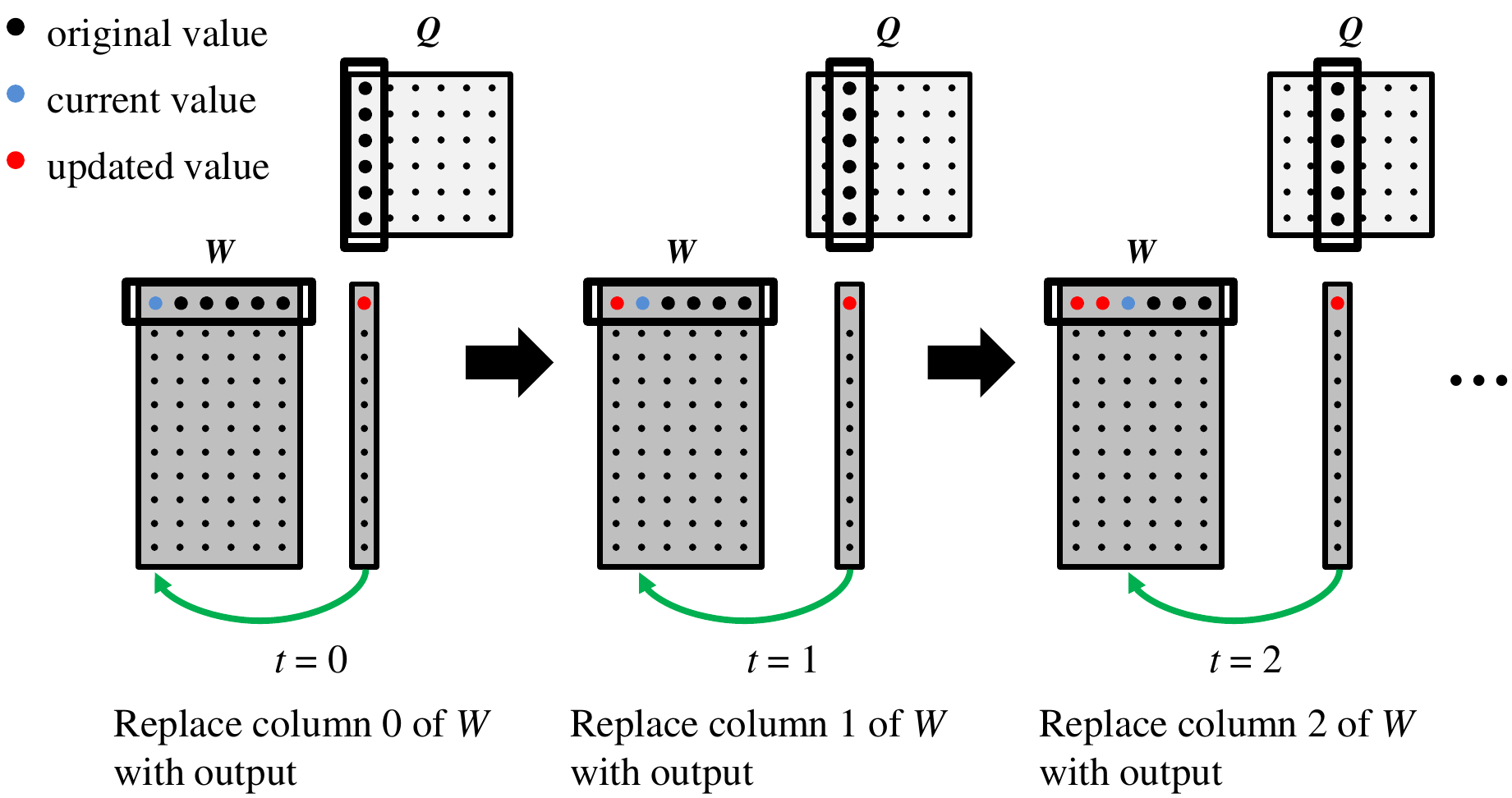}
	\caption{FAST-HALS: Update of $W$.}
	\label{fig:original_1}
\end{figure}

\begin{figure}
	\centering
	\includegraphics*[viewport= 20 250 550 530, clip=true, width=0.99\columnwidth]{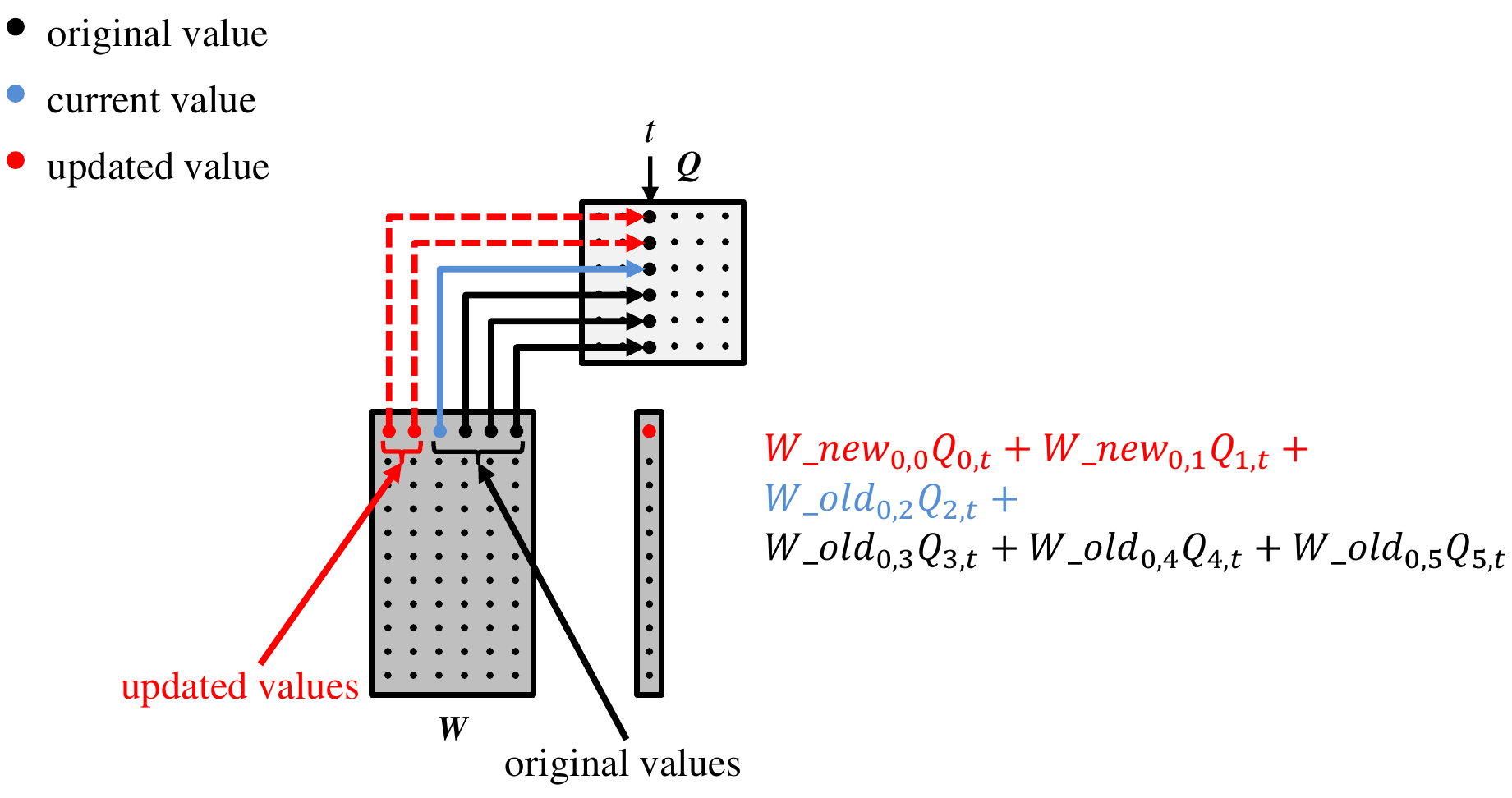}
	\caption{FAST-HALS: Updating a single element of $W$. The dash represents updated value.}
	\label{fig:original_2}
\end{figure}


In this sub-section, we describe how the FAST-HALS algorithm is adapted
by exploiting the flexibility of changing the order in which additive
contributions to a data element are made.
Before describing the adaptation, we first highlight
the interaction between different columns of $W$ in the original
algorithm. Figure \ref{fig:original_1}
depicts the update of $W$ which corresponds to the lines 12 to 16 in Algorithm
\ref{alg:FAST_HALS}. 

In Algorithm \ref{alg:FAST_HALS}, $t^{th}$ column of $W$ is updated as the
product of $W$ with $t^{th}$ column of $Q$ which is a matrix-vector
multiplication operation. Since the update to ${(t+1)}^{th}$ column depends on
$t^{th}$ column, different columns ($t$: features) are updated sequentially. Let
$W\_old$ represent the values at the beginning of the current outer iteration, and
let $W\_new$ represent the values at the end of current outer iteration (updated
values). Interaction between $W\_old$ and $W\_new$ is shown in Figure
\ref{fig:original_2} which depicts the contributions from $W\_old$ and $W\_new$
to $W\_new_{i,t}$.  $W\_new_{i,t}$ can be obtained by $\sum\limits_{j=0}^{t-1}
W\_new_{i,j} \times Q_{j,t} + \sum\limits_{j=t}^{K-1} W\_old_{i,j} \times
Q_{j,t}$.

\begin{figure}[h]
	\centering
	\includegraphics*[viewport= 20 350 450 530, clip=true, width=0.9\columnwidth]{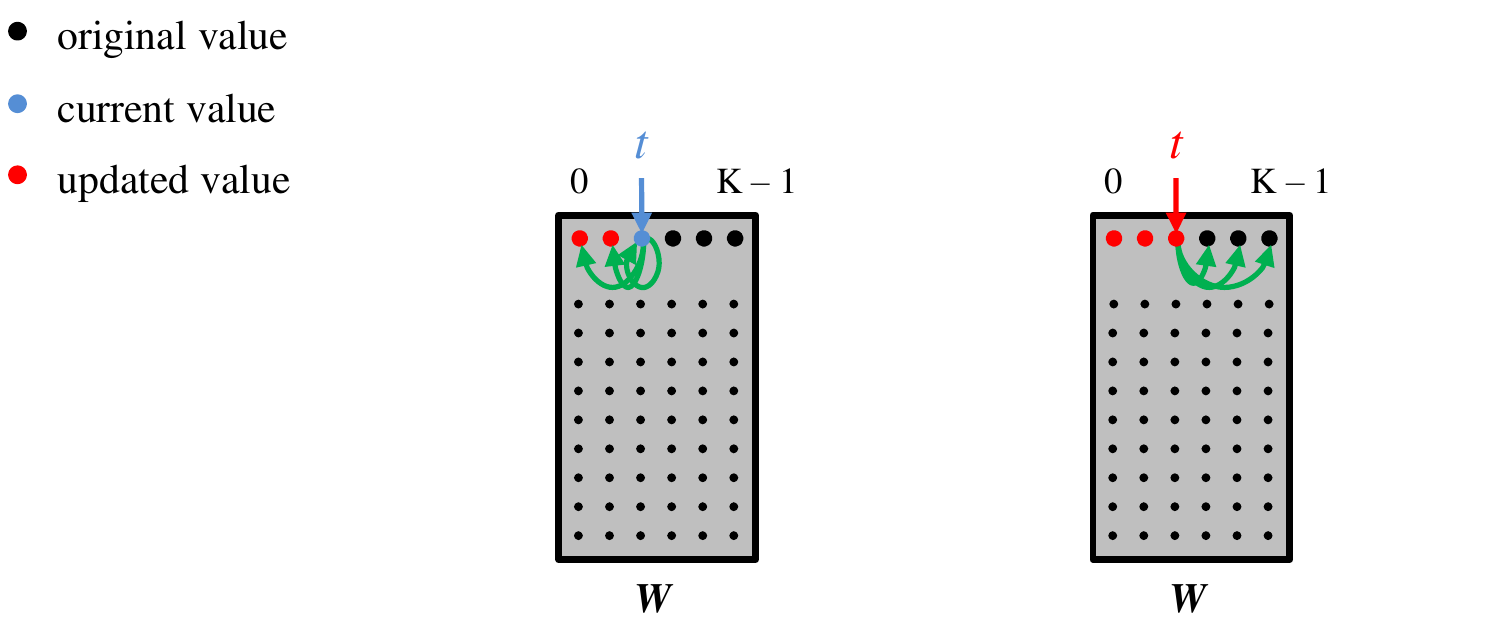}
	\caption{The contributions from $W_{0,t}$ to other elements.}
	\label{fig:original_3}
\end{figure}

Figure \ref{fig:original_3} shows the contributions of $W\_old_{i,t}$ and
$W\_new_{i,t}$ to $W\_new_{i,*}$. $W\_old_{i,t}$ contributes to $W\_new_{i,j}$
$\forall j | j\leq t$, and $W\_new_{i,t}$ contributes to $W\_new_{i,j}$ where
$\forall j| j > t$. In other words, the old value of column $t$ is used to
update the columns to the left of $t$ (and self), and the new/updated value of
column $t$ is used to update the columns to the right of column $t$.

\begin{figure}[h]
	\centering
	\includegraphics*[viewport= 5 290 580 530, clip=true, width=0.9\columnwidth]{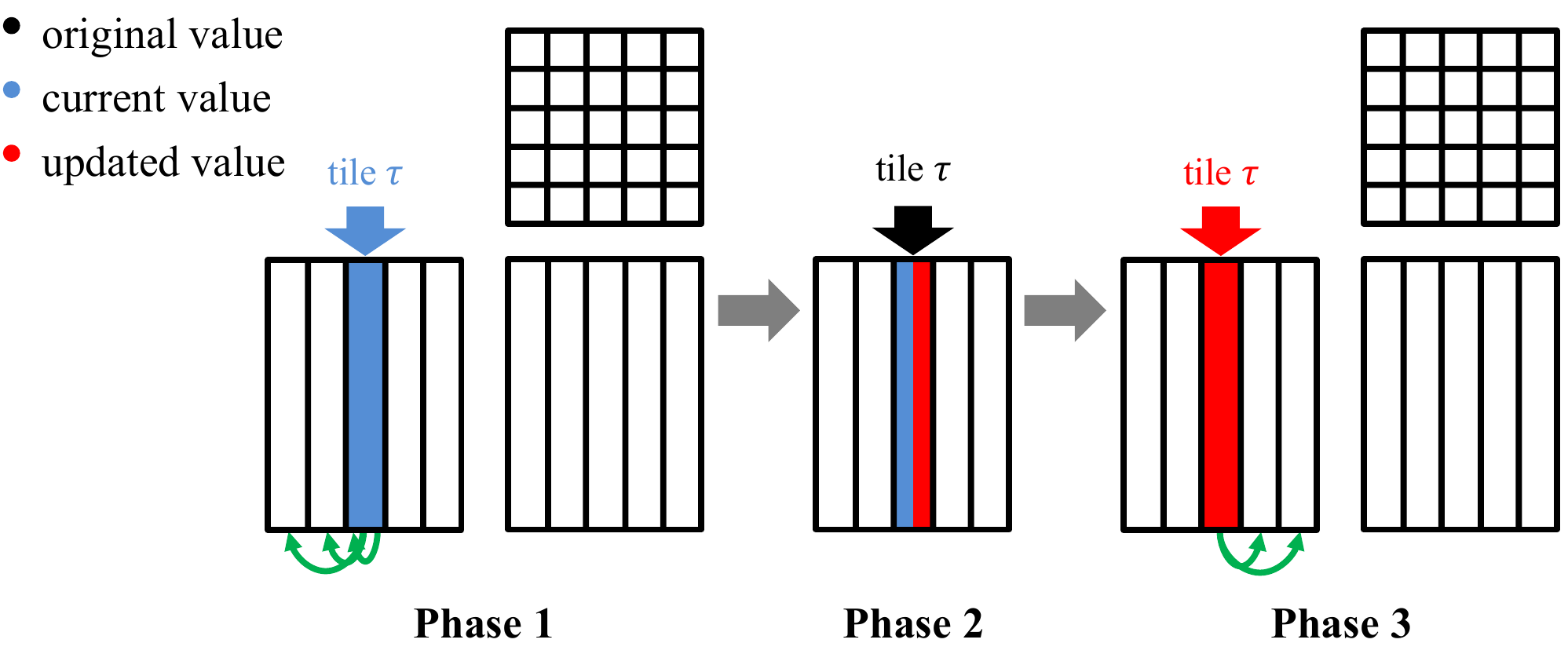}
	\caption{Overview of our approach for updating $W$.}
	\label{fig:alternate_overview}
\end{figure}

\begin{algorithm}
	\caption{Parallel CPU implementation for updating $W$}
	\label{alg:update_W_NMF}
	\begin{flushleft}
		\textbf{Input}: $A \in \mathbb R_+^{V \times D}$: input matrix, $W\_old$ and $W\_new$: $V \times K$ non-negative matrix factor, $H$: $D \times K$ non-negative matrix factor, $T$: Tile size, $\epsilon$: small non-negative quantity, $\gamma$: total number of tiles\\
		\end{flushleft}
		\begin{algorithmic}[1]
			\STATE $P$ $\leftarrow$ $AH^{T}$ 
			\STATE $Q$ $\leftarrow$ $HH^{T}$ 
			
			\STATE \textcolor{blue}{\textbf{// Initialize $W\_new$ using $W\_old$ and $Q$}}
			\FOR{v $=$ 0 \textbf{to} V $-$ 1}
				\FOR{k $=$ 0 \textbf{to} K $-$ 1}
					\STATE $W\_new$[v][k] $\leftarrow$ $W\_old$[v][k] $\times$ $Q$[k][k]
				\ENDFOR
			\ENDFOR
			\STATE \textcolor{blue}{\textbf{// Phase 1}}
			\STATE $\gamma$ $\leftarrow$ $K$ / $T$
			\FOR{tile\_id $=$ 0 \textbf{to} $\gamma$ $-$ 1}
			\STATE $W\_new$[0:$V$-1][0:(tile\_id$\times$ $T$)-1] $-=$ \\
			\textbf{dgemm}($W\_old$[0:$V$-1][tile\_id$\times~T$:((tile\_id +1)$\times$ $T$)-1], $Q$[tile\_id$\times$ $T$:((tile\_id +1)$\times$ $T$)-1][0:(tile\_id$\times$ $T$)-1])
			\ENDFOR
			\STATE \textcolor{blue}{\textbf{// Phase 2 \& Phase 3}}
			\FOR{tile\_id $=$ 0 \textbf{to} $\gamma$ $-$ 1}
				\STATE \textcolor{blue}{\textbf{// Phase 2}}
				\FOR{t $=$ tile\_id $\times$ $T$ \textbf{to} (tile\_id $+$ 1) $\times$ $T$ $-$ 1}
					\STATE sum\_square $\leftarrow$ 0
					\STATE \textbf{\#pragma omp parallel for reduction(+:sum\_square)}
					\FOR{v $=$ 0 \textbf{to} V $-$ 1}
						\STATE sum $\leftarrow$ 0
						\STATE k $\leftarrow$ tile\_id $\times$ $T$
						\STATE \textbf{\#pragma omp simd reduction(+:sum)}
						\FOR{; \textbf{to} t $-$ 1}
							\STATE sum $\leftarrow$ sum $+$ $W\_new$[v][k] $\times$ $Q$[t][k]	
						\ENDFOR
						\STATE \textbf{\#pragma omp simd reduction(+:sum)}
						\FOR{k $=$ t; \textbf{to} (tile\_id $+$ 1) $\times$ $T$ $-$ 1}
							\STATE sum $\leftarrow$ sum $+$ $W\_old$[v][k] $\times$ $Q$[t][k]	
						\ENDFOR
						\STATE $W\_new$[v][t] $\leftarrow$ max($\epsilon$, $W\_new$[v][t] $+$ $P$[v][t] $-$ sum)
						\STATE sum\_square $\leftarrow$ sum\_square $+$ $W\_new$[v][t] $\times$ $W\_new$[v][t]
					\ENDFOR
					\STATE \textbf{\#pragma omp parallel for}
					\FOR{v $=$ 0 \textbf{to} V $-$ 1}
						\STATE $W\_new$[v][t] $\leftarrow$ $W\_new$[v][t] $/$ sqrt(sum\_square)
					\ENDFOR
				\ENDFOR
				\STATE \textcolor{blue}{\textbf{// Phase 3}}
				\STATE $W\_new$[0:$V$-1][(tile\_id +1)$\times$ $T$:$K$-1] $-=$ \\
				\textbf{dgemm}($W\_new$[0:$V$-1][tile\_id$\times$ $T$:((tile\_id +1)$\times$ $T$)-1], $Q$[tile\_id$\times$ $T$:((tile\_id +1)$\times$ $T$)-1][(tile\_id +1)$\times$ $T$:$K$-1])
			\ENDFOR
		\end{algorithmic}
\end{algorithm}

If we partition $W$ into a set of column panels (tiles) of size $T$, the
interactions between columns can be expressed in terms of tiles as depicted in
Figure \ref{fig:alternate_overview}. Similar to individual columns, the old
value of tile $\tau$ is used to update the columns to the left of $\tau$ (phase 1), and
the new/updated value of tile $\tau$ is used to update the tiles to the right of
tile $\tau$ (phase 3). The updates to different columns with a tile (phase 2) is
done sequentially.

The contributions to tiles to the left of current tile $\tau$
can be done as $W\_new_{i,j}-=W\_old_{i,\tau \times T:((\tau+1) \times T)-1} \times
Q_{\tau \times T:((\tau+1) \times T)-1,j}$ where $\forall j | j < \tau \times T-1$. Similarly, contributions to tiles to the right of current tile $\tau$ can be done as $W\_new_{i,j}-=W\_new_{i,\tau \times T:((\tau+1) \times T)-1} \times Q_{\tau \times T:((\tau+1) \times T)-1,j}$ where $\forall j | j > (\tau+1) \times T$. Both phases 1 and 3 can be performed using matrix-matrix operations which are known to have much better performance and lower data movement than matrix-vector operations. Note that the total number of operations in both the original formulation and our formulation are exactly the same.

\begin{figure}[h]
	\centering
	\includegraphics*[viewport= 5 260 720 540, clip=true, width=1\columnwidth]{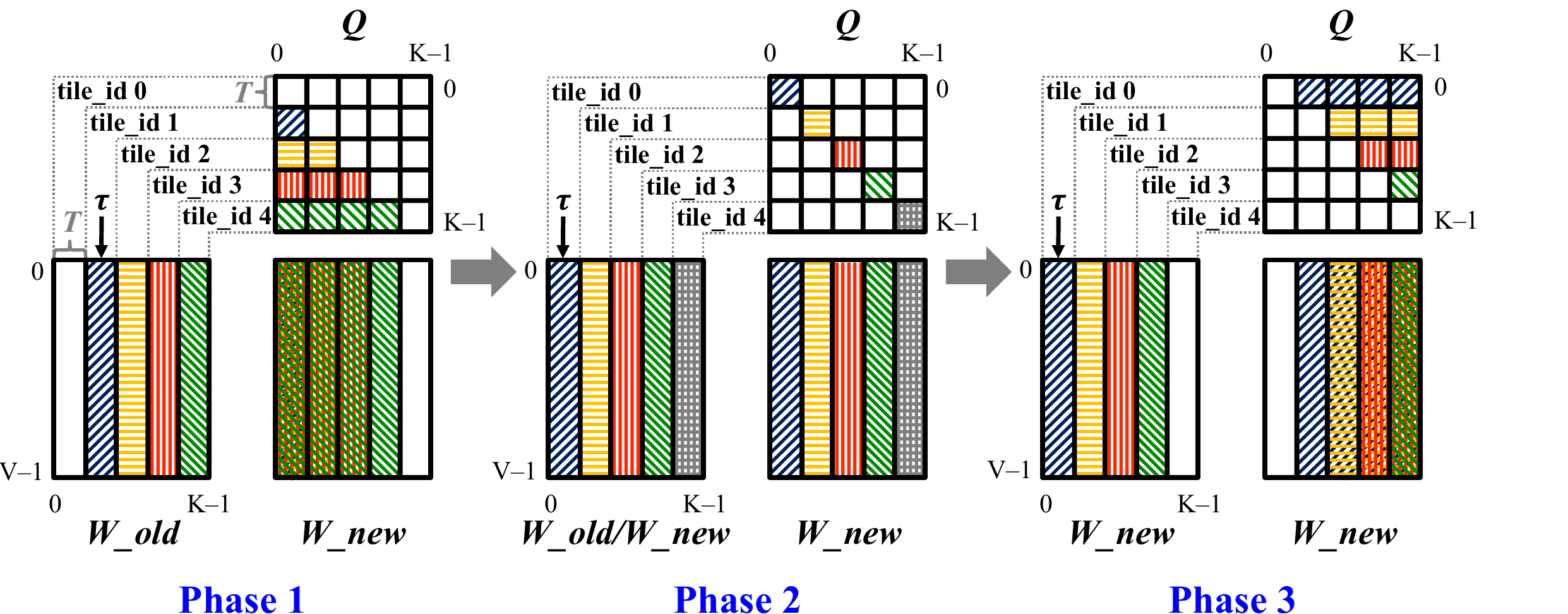}
	\caption{Computations of three phases for updating $W$.}
	\label{fig:alternate_detail}
\end{figure}




\section{Details of PL-NMF on Multicore CPUs and GPUs}
\subsection{Parallel CPU Implementation}
Algorithm \ref{alg:update_W_NMF} shows our CPU pseudo-code for updating $W$. We
begin by computing $AH^T$ (line 1). If $A$ is sparse, then the actual
implementation uses \textbf{mkl\_dcsrmm}() and \textbf{cblas\_dgemm}()
is used otherwise. Line 2 computes the $HH^T$ (using \textbf{cblas\_dgemm}()).
The $W$ values from the previous iteration are kept in $W\_old$. We maintain another
data structure called $W\_new$ which represents the updated $W$ values. $W\_new$
is initialized by the loop in line 4. By using Equation \ref{eq:phase_1}, phase 1 is done by the loop in line 11. Figure \ref{fig:alternate_detail} illustrates the actual computations of tiled matrix-matrix multiplications for three sequential phases, where $\tau$ denotes the index of the current tile and $T$ is the size of each tile. For example, at current tile $\tau$, phase 1 performs multiplication of the same colored/patterned two sub-matrices (tiles) in $W\_old$ and $Q$ to update the result matrix $W\_new$.
\begin{equation} \label{eq:phase_1}
\footnotesize
\begin{multlined}
W\_new[\,:\,,\,0:(\tau \times T)-1]\,-= \\
W\_old[\,:\,,\,(\tau \times T):((\tau+1) \times T)-1] \cdot \\
Q[(\tau \times T):((\tau+1) \times T)-1,\,0:(\tau \times T)-1]
\end{multlined}
\end{equation}
The loop in line 17 performs phase 2 computations as formulated in Equation \ref{eq:phase_2}. In order to take advantage of the vector units, the loops in lines 24 and 28 are vectorized. Additionally, a column-wise normalization for $W\_new$ is performed within phase 2 (line 36).
\begin{equation} \label{eq:phase_2}
\footnotesize
\begin{multlined}
W\_new[\,:\,,\,(\tau \times T):((\tau+1) \times T)-1]\,-= \\
W[\,:\,,\,(\tau \times T):((\tau+1) \times T)-1] \cdot \\
Q[(\tau \times T):((\tau+1) \times T)-1,\,(\tau \times T):((\tau+1) \times T)-1]\\
+P[\,:\,,\,(\tau \times T):((\tau+1) \times T)-1]
\end{multlined}
\end{equation}
The matrix-matrix multiplication in line 40 corresponds to the phase 3 computations using Equation \ref{eq:phase_3}. As depicted in Figure \ref{fig:alternate_detail}, the tiles involving phase 3 and phase 1 computations are different from each other.
\begin{equation} \label{eq:phase_3}
\footnotesize
\begin{multlined}
W\_new[\,:\,,\,((\tau+1) \times T):K-1]\,-= \\
W\_new[\,:\,,\,(\tau \times T):((\tau+1) \times T)-1] \cdot \\
Q[(\tau \times T):((\tau+1) \times T)-1,\,((\tau+1) \times T):K-1]
\end{multlined}
\end{equation}
Finally, our parallel CPU implementation completely substitutes lines 10 to 16 in Algorithm \ref{alg:FAST_HALS} for all lines in Algorithm \ref{alg:update_W_NMF}. Similarly, $H$ will be updated in the same fashion as updating $W$ except for the normalization part.

\subsection{Parallel GPU Implementation}
\begin{algorithm}
	\caption{GPU implementation of updating W on host}
	\label{alg:update_W_NMF_host}
	\begin{flushleft}
		\textbf{Input}: $A \in \mathbb R_+^{V \times D}$: input matrix, $W\_old$ and $W\_new$: $V \times K$ non-negative matrix factor, $H$: $D \times K$ non-negative matrix factor, $T$: Tile size, $\epsilon$: small non-negative quantity, $\gamma$: total number of tiles\\
		\end{flushleft}
		\begin{algorithmic}[1]
			\STATE $P$ $\leftarrow$ $AH^{T}$ 
			\STATE $Q$ $\leftarrow$ $HH^{T}$ 
			\STATE \textcolor{blue}{\textbf{// Initialize $W\_new$ using $W\_old$ and $Q$}}
			\STATE init\_W\_new()
			\STATE \textcolor{blue}{\textbf{// Phase 1}}
			\STATE $\gamma$ $\leftarrow$ $K$ / $T$
			\FOR{tile\_id $=$ 0 \textbf{to} $\gamma$ $-$ 1}
				\STATE $W\_new$[0:$V$-1][0:(tile\_id$\times$ $T$)-1] $-=$ \\
			\textbf{cublasDgemm}($W\_old$[0:$V$-1][tile\_id$\times$ $T$:((tile\_id +1)$\times$ $T$)-1], $Q$[tile\_id$\times$ $T$:((tile\_id +1)$\times$ $T$)-1][0:(tile\_id$\times$ $T$)-1])
			\ENDFOR
			\STATE \textcolor{blue}{\textbf{// Phase 2 \& Phase 3}}
			\FOR{tile\_id $=$ 0 \textbf{to} $\gamma$ $-$ 1}
				\STATE \textcolor{blue}{\textbf{// Phase 2}}
				\FOR{t $=$ tile\_id $\times$ $T$ \textbf{to} (tile\_id $+$ 1) $\times$ $T$ $-$ 1}
					\STATE \textbf{cudaMemset}($sum\_square$,0)
					\STATE update\_W\_phase\_2()
					\STATE \_\_cudaDeviceSynchronize()
					\STATE update\_W\_norm()
					\STATE \_\_cudaDeviceSynchronize()	
				\ENDFOR
				\STATE \textcolor{blue}{\textbf{// Phase 3}}
				\STATE $W\_new$[0:$V$-1][(tile\_id +1)$\times$ $T$:$K$-1] $-=$ \\
				\textbf{cublasDgemm}($W\_new$[0:$V$-1][tile\_id$\times$ $T$:((tile\_id +1)$\times$ $T$)-1], $Q$[tile\_id$\times$ $T$:((tile\_id +1)$\times$ $T$)-1][(tile\_id +1)$\times$ $T$:$K$-1])
			\ENDFOR
		\end{algorithmic}
\end{algorithm}

\begin{algorithm}
	\caption{GPU implementation of update\_W\_phase\_2 kernel}
	\label{alg:update_W_NMF_device}
	\begin{flushleft}
		\textbf{Input}: $W\_old$, $W\_new$, $P$, $Q$, $sum\_square$, t, tile\_id, $T$, $V$, $K$, $\epsilon$\\
		\end{flushleft}
		\begin{algorithmic}[1]
			\STATE vId $\leftarrow$ blockIdx.x $\times$ blockDim.x $+$ threadIdx.x // threadID \\
			\STATE \_\_shared\_\_ $shared\_sum$[1024/32] \\
			\STATE sum\_reduce = 0.0f \\
			
			\IF{vId $<$ $V$}
				\STATE sum = 0 \\
				\FOR{k $=$ tile\_id $\times$ $T$ \textbf{to} (tile\_id $+$ 1) $\times$ $T$ $-$ 1}
					\IF{k $<$ t}
						\STATE sum $\leftarrow$ sum + $W\_new$[vId $+$ k $\times$ $V$][k] $\times$ $Q$[k $\times$ $K$ + t]
					\ELSE
						\STATE sum $\leftarrow$ sum + $W\_old$[vId $+$ k $\times$ $V$] $\times$ $Q$[k $\times$ $K$ + t]
					\ENDIF
				\ENDFOR
				\STATE $W\_new$[vId $+$ t $\times$ $V$] $\leftarrow$ max($\epsilon$, $W\_new$[vId $+$ t $\times$ $V$] $+$ $P$[vId $+$ t $\times$ $V$] $-$ sum) \\
				\STATE sum\_reduce $\leftarrow$ $W\_new$[vId $+$ t $\times$ $V$] \\
			\ENDIF
			
			\STATE sum\_reduce $\leftarrow$ sum\_reduce $\times$ sum\_reduce
			\STATE \textcolor{black}{\textbf{// Warp-level reduction}}
			
			\STATE sum\_reduce $\leftarrow$ \textbf{warp\_reduce}(sum\_reduce) \\
			
			\STATE \textcolor{black}{\textbf{// Block-level reduction}}
			\IF{threadIdx.x \% 32 == 0}
				\STATE $shared\_sum$[threadIdx.x / 32] $\leftarrow$ sum\_reduce
			\ENDIF
			\STATE \_\_syncthreads()
			\IF{threadIdx.x / 32 == 0}
				\STATE sum\_reduce $\leftarrow$ $shared\_sum$[threadIdx.x]
				\STATE sum\_reduce $\leftarrow$ \textbf{warp\_reduce}(sum\_reduce) \\
			\ENDIF
			
			\IF{threadIdx.x == 0}
				\STATE \textbf{atomicAdd} ($sum\_square$, sum\_reduce)
			\ENDIF

		\end{algorithmic}
\end{algorithm}

\begin{algorithm}
	\caption{GPU implementation of update\_W\_norm kernel}
	\label{alg:norm_W_NMF_device}
	\begin{flushleft}
		\textbf{Input}: $W\_new$, $sum\_square$, t, $V$\\
		\end{flushleft}
		\begin{algorithmic}[1]
			\STATE vId $\leftarrow$ blockIdx.x $\times$ blockDim.x $+$ threadIdx.x // threadID \\
			\IF{vId $<$ $V$}
				\STATE return
			\ENDIF
			\STATE $W\_new$[vId $+$ t $\times$ $V$] $\leftarrow$ $W\_new$[vId $+$ t $\times$ $V$] / sqrt($sum\_square$)
		\end{algorithmic}
\end{algorithm}
Similar to our CPU algorithm, our GPU algorithm also tries to minimize the data
movement. Algorithm \ref{alg:update_W_NMF_host}, \ref{alg:update_W_NMF_device}
and \ref{alg:norm_W_NMF_device} show the pseudo-code of our GPU algorithm.
Since the overall structure of the GPU algorithm is similar to the CPU
algorithm, this section only highlights the differences. Algorithm
\ref{alg:update_W_NMF_host} runs on the host which is responsible for launching GPU
kernels. The sparse matrix-dense matrix multiplication is implemented using
\textbf{cusparseDcsrmm()}, and dense matrix-dense matrix multiplication is
implemented using \textbf{cublasDgemm()}. 

Algorithm \ref{alg:update_W_NMF_device} shows the pseudo-code for phase 2. In
GPUs, the reduction across $V$ (for normalization of $W$) can be performed using
global memory atomic operations which are very expensive. Hence, our
implementation uses efficient hierarchical reduction. The reduction within a
thread block is done in 4 steps: i) in line 18, the reduction across the threads
within a warp is done using efficient warp shuffling primitives, ii) all the
threads with lane id 0 write the reduced value to shared memory (line 20), iii)
in line 24, the first warp of the thread block loads the previously written
values from shared memory and iv) all the threads in the first warp again performs
warp reduction (line 26).  In order to perform reduction across multiple thread
blocks, we use atomic operations which is shown in line 29. Algorithm
\ref{alg:norm_W_NMF_device} shows the pseudo-code for normalization.

\begin{figure*}[h!]
\centering
  \includegraphics[width=0.21\linewidth]{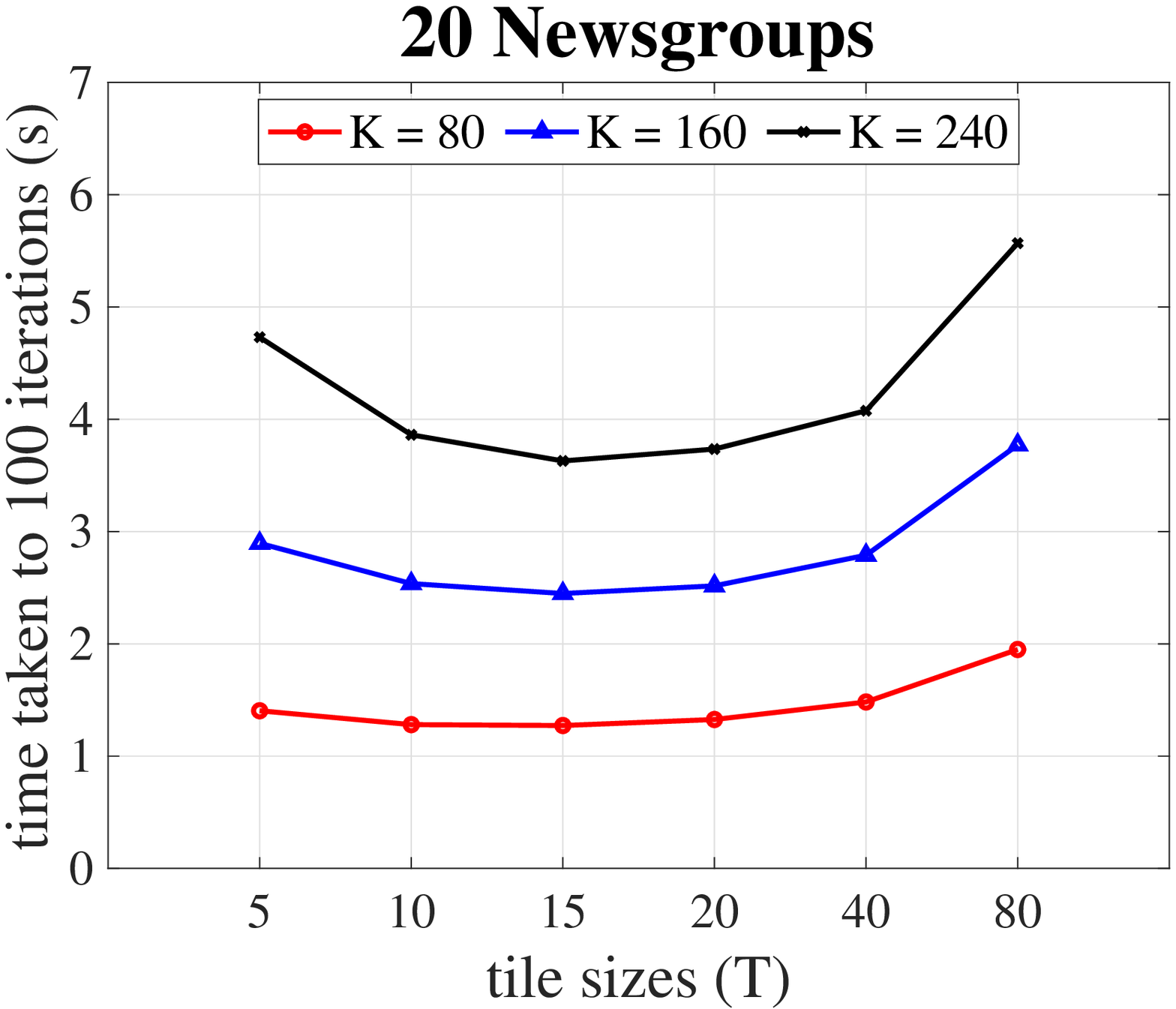}
  \hspace{-0.4cm}
  \includegraphics[width=0.21\linewidth]{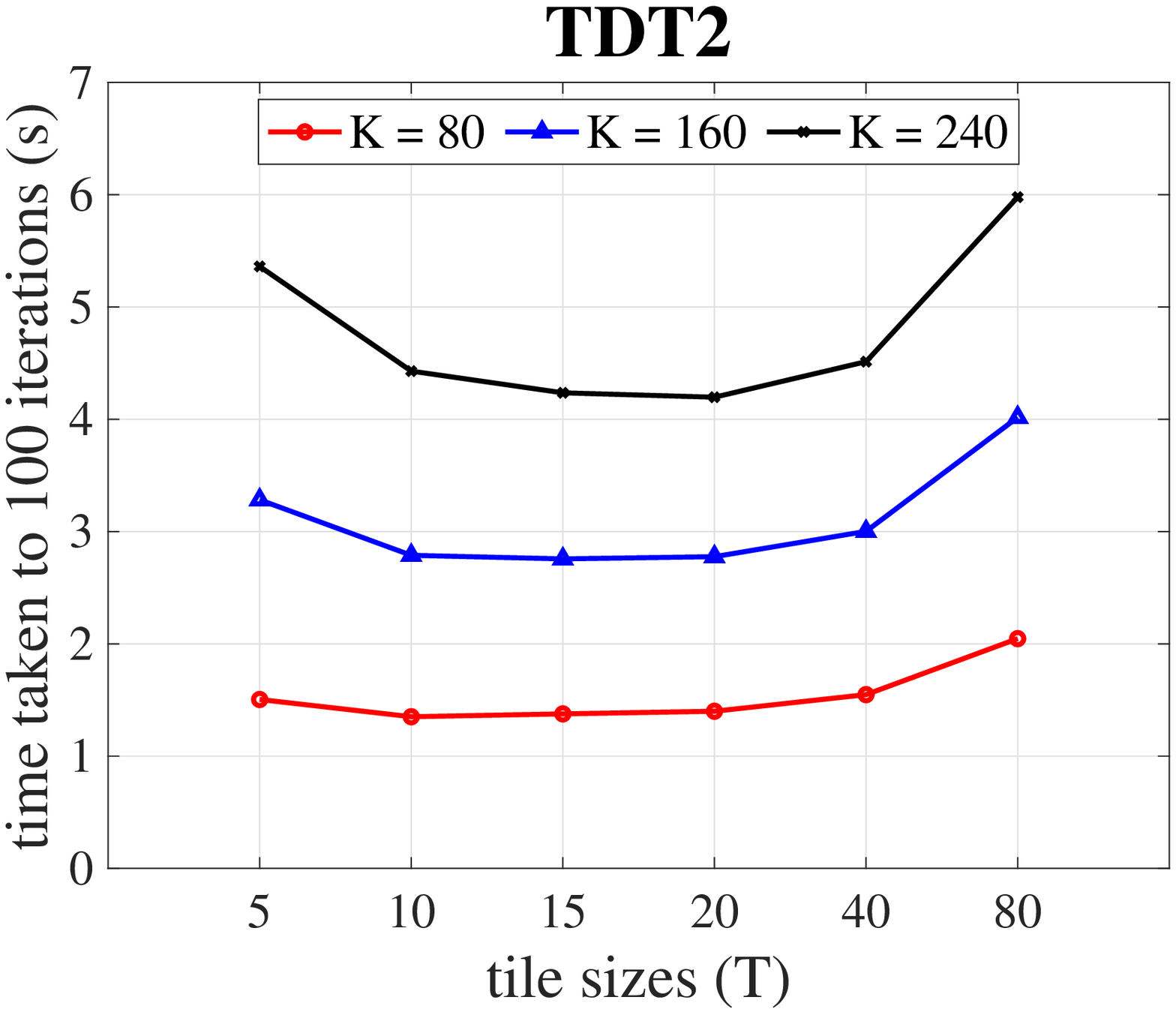}
  \hspace{-0.4cm}
  \includegraphics[width=0.21\linewidth]{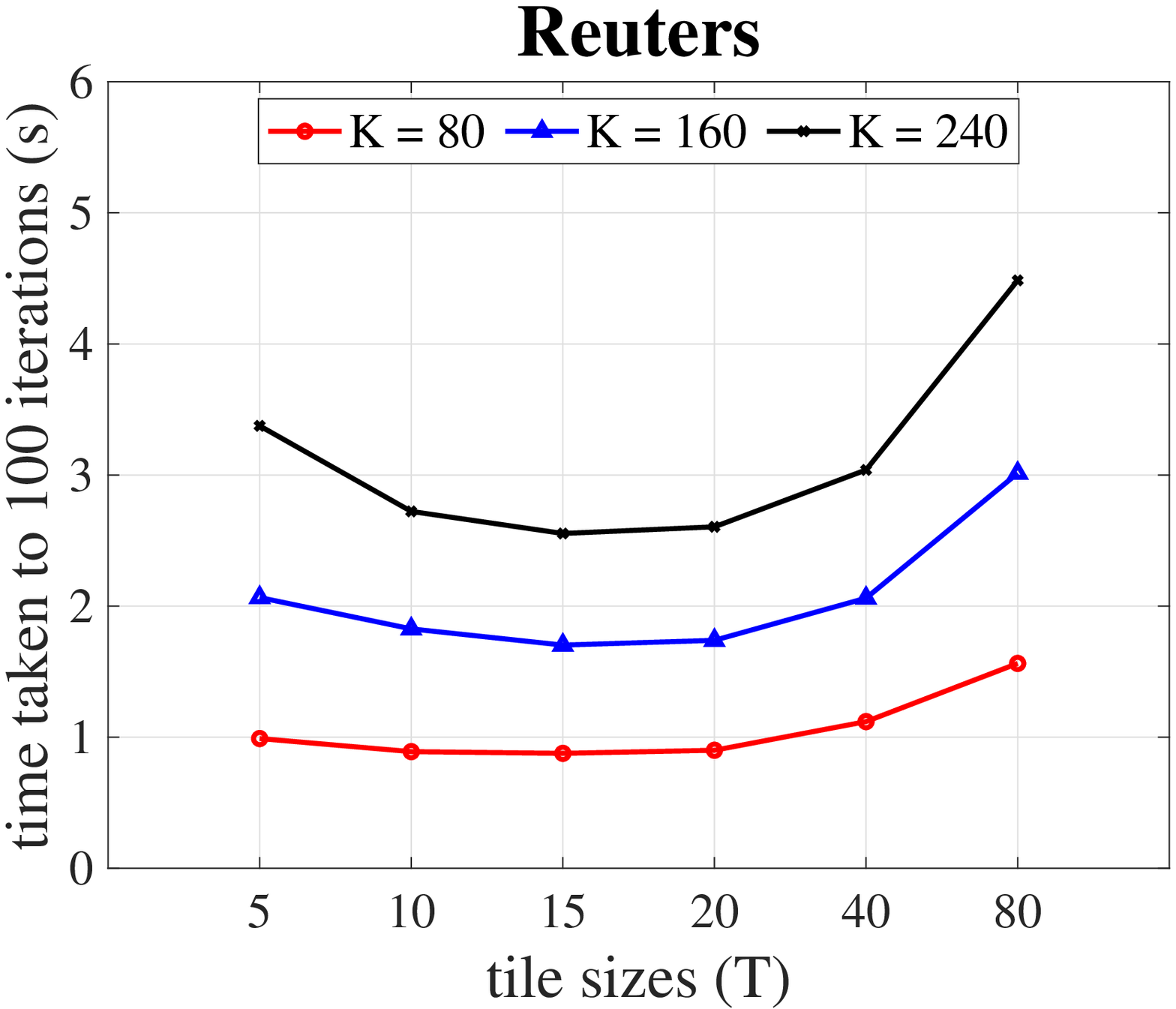}
  \hspace{-0.4cm}
  \includegraphics[width=0.21\linewidth]{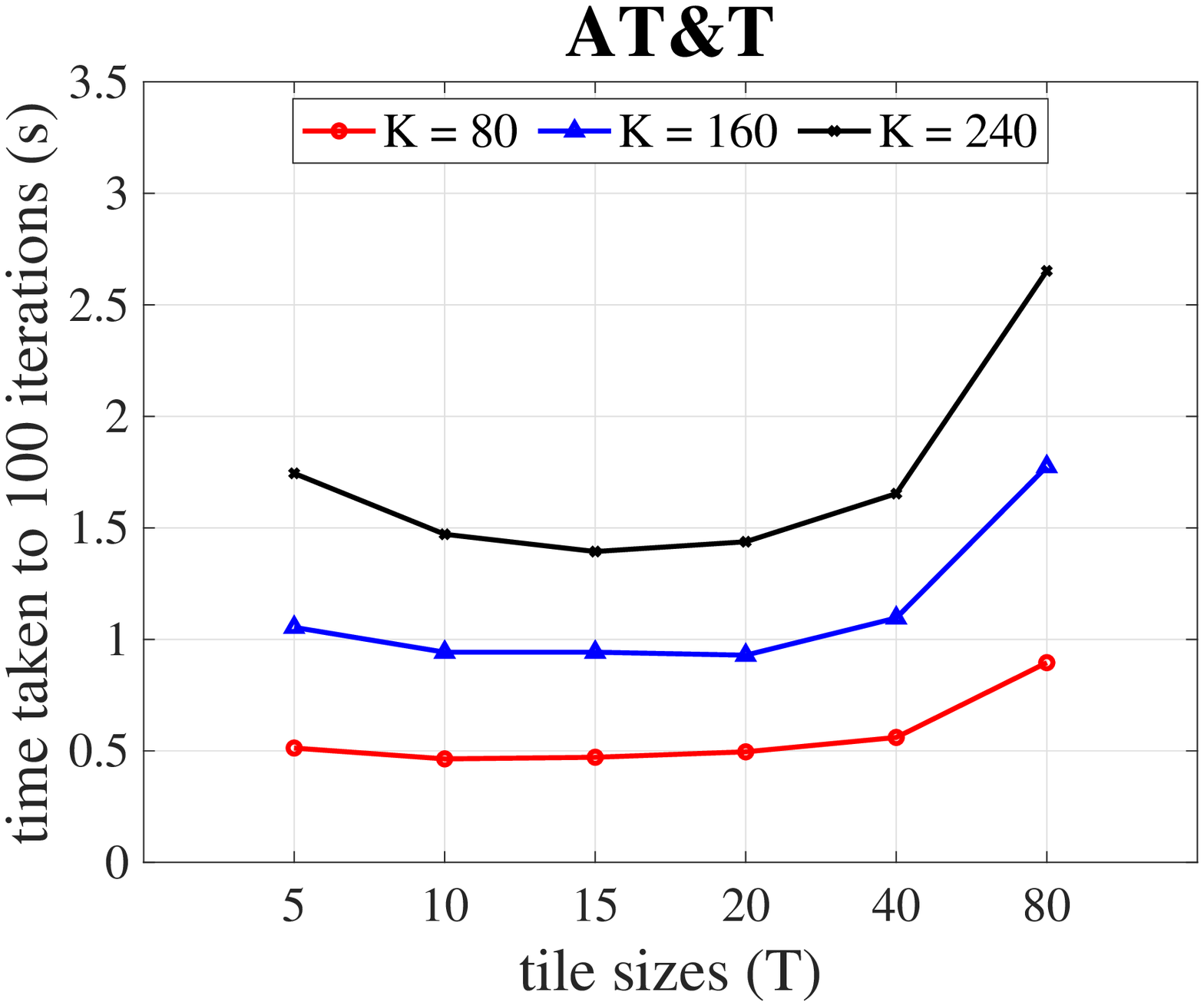}
  \hspace{-0.4cm}
  \includegraphics[width=0.21\linewidth]{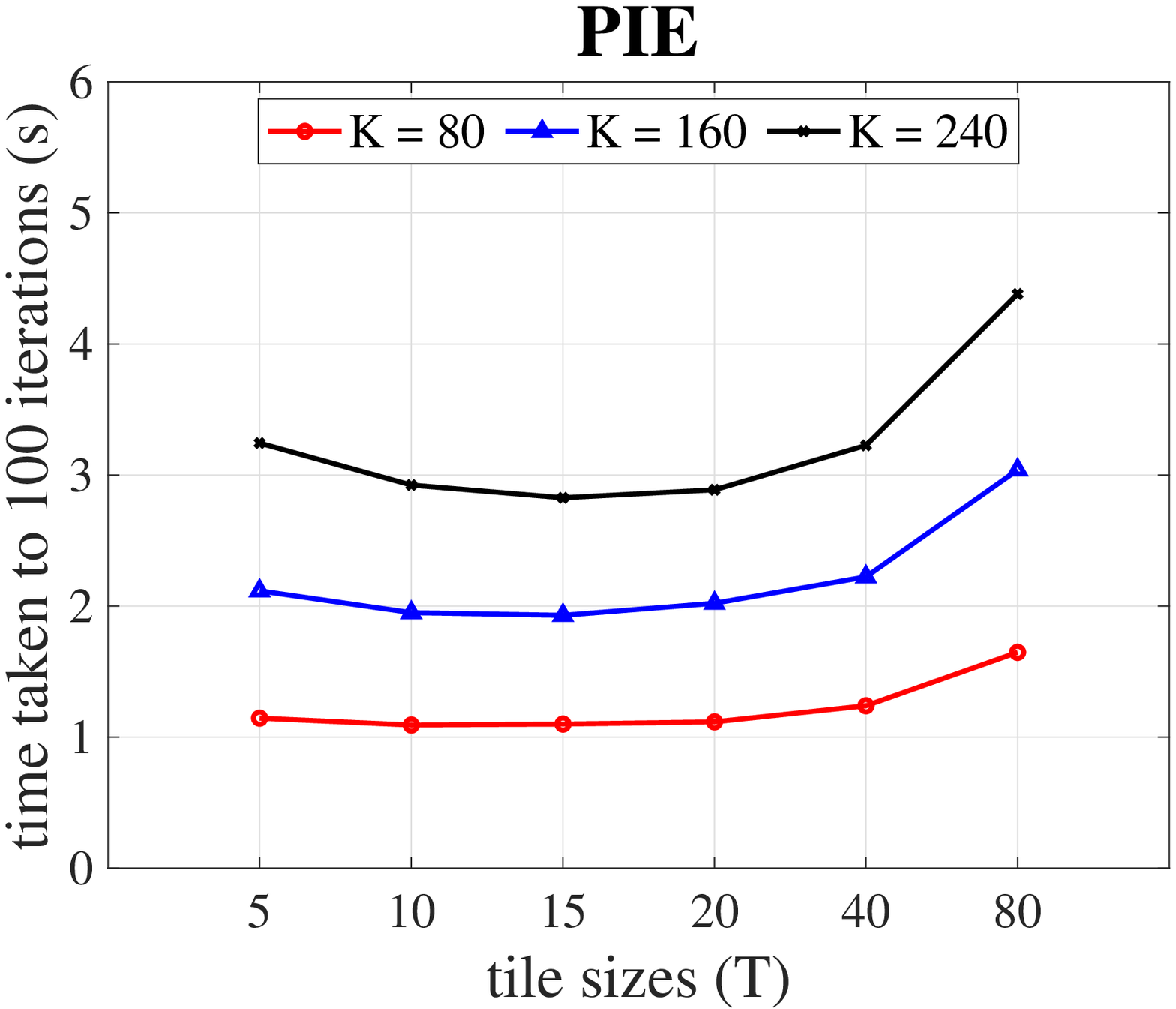}
  \caption{The time taken to reach 100 iterations when the tile size $T$ is
  varied for different $K$ on five datasets. Low-rank $K$ is set to 80, 160 and 240,
  and $T$ is varied for each $K$. X-axis: tile size; Y-axis: elapsed time to
  reach 100 iterations.}
  ~\label{fig:nmf_tile_time} 
\end{figure*}

\section{Modeling: Determination of the tile size}

In this section, we first compare the data movement cost of our approach with original FAST-HALS algorithm. Then the data movement of our algorithm as a function of $T$ is developed to select effective tile sizes.
\begin{equation}
	\label{eq:phase1_data_movement}
	\sum_{i=0}^{\frac{K}{T}-1}iVT^{2}(\frac{1}{T} + \frac{2}{\sqrt{C}}) = VT^{2}(\frac{1}{T}+\frac{2}{\sqrt{C}})(\frac{K^{2}-KT}{2T^{2}})
\end{equation}
\begin{equation}
	\label{eq:phase2_data_movement}
	\sum_{i=0}^{\frac{K}{T}-1}T(VT+T+V) = \frac{K}{T}T(VT+T+V)
\end{equation}
In our approach, $W$ is updated in three phases. Phases 1 and 3 can be implemented using matrix-multiplication, and the
corresponding cost is shown in Equation \ref{eq:phase1_data_movement}, where $T$
represents the tile size and $C$ is the cache size. Phase 2 can be implemented
using matrix-vector multiplication and the associated cost is shown in Equation
\ref{eq:phase2_data_movement}.  Since loading matrix $W$ dominates the data
movement cost in phase 2, the cost of loading vectors can be ignored.  Equation
\ref{eq:tNMF_data_movement} shows the total data movement required for updating
$W$.

\begin{equation}
    \label{eq:tNMF_data_movement}
    vol(T) = V(\frac{1}{T}+\frac{2}{\sqrt{C}})(K^{2}-KT) + \frac{K}{T}T(VT)
\end{equation}
The cost of updating $H$ is similar to updating $W$. Compared to updating $W$,
updating $H$ does not require accessing $Q$. In addition, since $H$ is not
normalized, the cost associated with normalization is also not present.

The data movement cost of the original loop in line 12 in Algorithm
\ref{alg:FAST_HALS} is $K(VK + K + 6V + 1)$. Hence, for the 20 Newsgroups
dataset ($V$=11,314) with low rank $K$=160 on a machine with 35 MB cache, the
data movement cost of original scheme is 300,525,600. However, 
in our scheme based on Equation \ref{eq:tNMF_data_movement}, the cost is only 44,897,687 which is 6.7$\times$ lower than the original scheme.



The tile size $T$ affects the data movement volume and hence the performance. Equation \ref{eq:tNMF_data_movement} shows the data movement of our algorithm as a function of $T$. Consider the case when there is only one tile ($T = K$). In this case, there is no work associated with phase 1 (contributions to left) and phase 3 (contributions to the right) as the first term of Equation
\ref{eq:tNMF_data_movement} will become zero. The total data movement of phase 2 is $VK^2$ which is very high.  Now consider the other extreme where the tile size is 1 ($T=1$). In this case, phases 1 and 2 have very high data movement ($ >VK^2$). Thus, when $T$ is high, the total data movements required for phases 1 and
3 are low, but phase 2 has high data movement. On the other extreme, when $T$ is
low, the total data movements for phases 1 and 3 are high, but phase 2 has low data
movement. Hence, we expect the combined data movement for all the phases to
decrease as $T$ increases from 1 to some point and then the data movement will
increase again as $T$ approaches $K$. Since performance is correlated with data
movement, the performance as a function of tile size should show a similar trend
and is shown in Figure \ref{fig:nmf_tile_time}.

\begin{equation}
	\label{eq:derivative_model}
	\frac{d(vol(T))}{dT} = T^{2}(\frac{2}{\sqrt{C}}-1) + K = 0
\end{equation}
\begin{equation}
	\label{eq:final_derivative_model}
	T = \sqrt{K-\frac{2}{\sqrt{C}}}
\end{equation}

In order to build a model to determine the tile size for a given $K$, the
derivative of Equation \ref{eq:tNMF_data_movement} with respect to $T$ is set it
to zero (Equation \ref{eq:derivative_model}). The solution to Equation
\ref{eq:derivative_model} is shown in Equation \ref{eq:final_derivative_model}.
For a machine with cache size of 35 MB, the tile sizes computed by our model are
8.94, 12.64 and 15.49 for K=80, 160 and 240, respectively. Figure
\ref{fig:nmf_tile_time} shows that our model selected tile sizes that are
optimal/near optimal.
\section{Experimental Evaluation}
\label{section:experiments}
This section compares the time to convergence and convergence rate of PL-NMF with various state-of-the-art techniques.

\subsection{Benchmarking Machines}
Table \ref{tb:machine_nmf} shows the configuration of the benchmarking machines
used for experiments.  All the CPU experiments were run on an Intel Xeon CPU
E5-2680 v4 running at 2.4 GHz with 128GB RAM. The GPU experiments were run on an
NVIDIA Tesla P100 PCIE GPU with 16GB global memory.
\begin{table}[h]
\centering
\caption{Machine configuration}
\label{tb:machine_nmf}
\scalebox{0.8}{
\begin{tabular}{|c|c|}
\hline
\textbf{Machine} & \textbf{Details}                                                                                                                         \\ \hline
CPU              & \begin{tabular}[c]{@{}c@{}}Intel(R) Xeon(R) CPU E5-2680 v4 (28 cores), 128GB\\ ICC version 18.0.3\end{tabular}                           \\ \hline
GPU              & \begin{tabular}[c]{@{}c@{}}Tesla P100 PCIE\\ (56 SMs, 64 cores/MP, 16GB Global Memory, 4 MB L2 cache)\\ CUDA version 9.2.88\end{tabular} \\ \hline
\end{tabular}}
\end{table}

\subsection{Datasets}

For experimental evaluations we used three publicly available real-world text datasets --
20~Newsgroups\footnote{\scriptsize{http://dengcai.zjulearning.org:8081/Data/TextData.html}\label{textdata}},
TDT2\footref{textdata}, Reuters\footref{textdata}. In addition, in order to
represent the audio-visual context analysis in social media platforms, we used
two image datasets --
AT\&T\footnote{\scriptsize{https://www.cl.cam.ac.uk/research/dtg/attarchive/facedatabase.html}}
and PIE\footnote{\scriptsize{http://dengcai.zjulearning.org:8081/Data/FaceDataPIE.html}}.
20~Newsgroups, TDT2 and Reuters are sparse matrices, and AT\&T and PIE are dense
matrices. The 20~Newsgroups dataset contains a document-term matrix in bag-of-words
representation associated with 20 topics. TDT2 (Topic Detection and Tracking 2)
dataset is a collection of text documents from CNN, ABC, NYT, APW, VOA and PRI.
Reuters dataset is a collection of documents from the Reuters newswire in 1987.
Both AT\&T and PIE datasets contain images of faces in dense matrix format. The
size of each image in AT\&T and PIE datasets is 92$\times$112 and 64$\times$64
pixels, respectively. Table \ref{tb:dataset_nmf} shows the characteristics of
each dataset.

\begin{table}[h]
\centering
\caption{Statistics of datasets used in the experiments. $V$ is the number of rows and $D$ is the number of columns in non-negative matrix $A$. For the text datasets, $V$ is the vocabulary size and $D$ is the number of documents.}
\label{tb:dataset_nmf}
\scalebox{0.8}{
\begin{tabular}{|c|c|c|c|c|}
\hline
\textbf{Dataset} & \textbf{\textit{V}} & \textbf{\textit{D}} & \textbf{Total NNZ} & \textbf{Sparsity (\%)} \\ \hline
20 Newsgroups    & 26,214       & 11,314       & 1,018,191          & 99.6567                \\ \hline
TDT2             & 36,771       & 10,212       & 1,323,869          & 99.6474                \\ \hline
Reuters          & 18,933       & 8,293        & 389,455            & 99.7519                \\ \hline
AT\&T            & 400          & 10,304       & 4,121,478          & 0.0030                 \\ \hline
PIE              & 11,554       & 4,096        & 47,321,408         & 0.0080                 \\ \hline
\end{tabular}
}
\end{table}

\begin{figure*}[h!]
  \centering
  \includegraphics[width=0.3\linewidth]{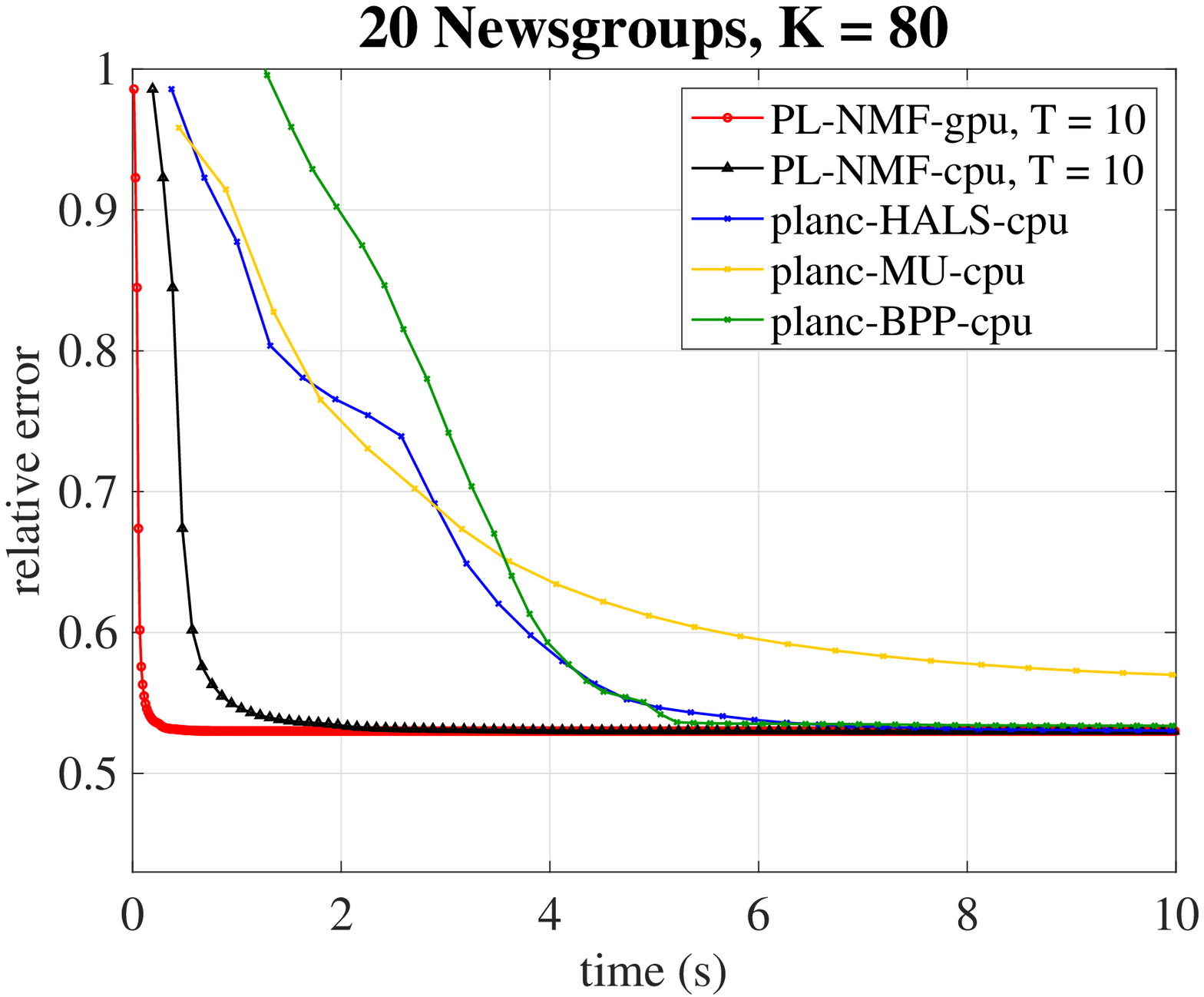}
  \includegraphics[width=0.3\linewidth]{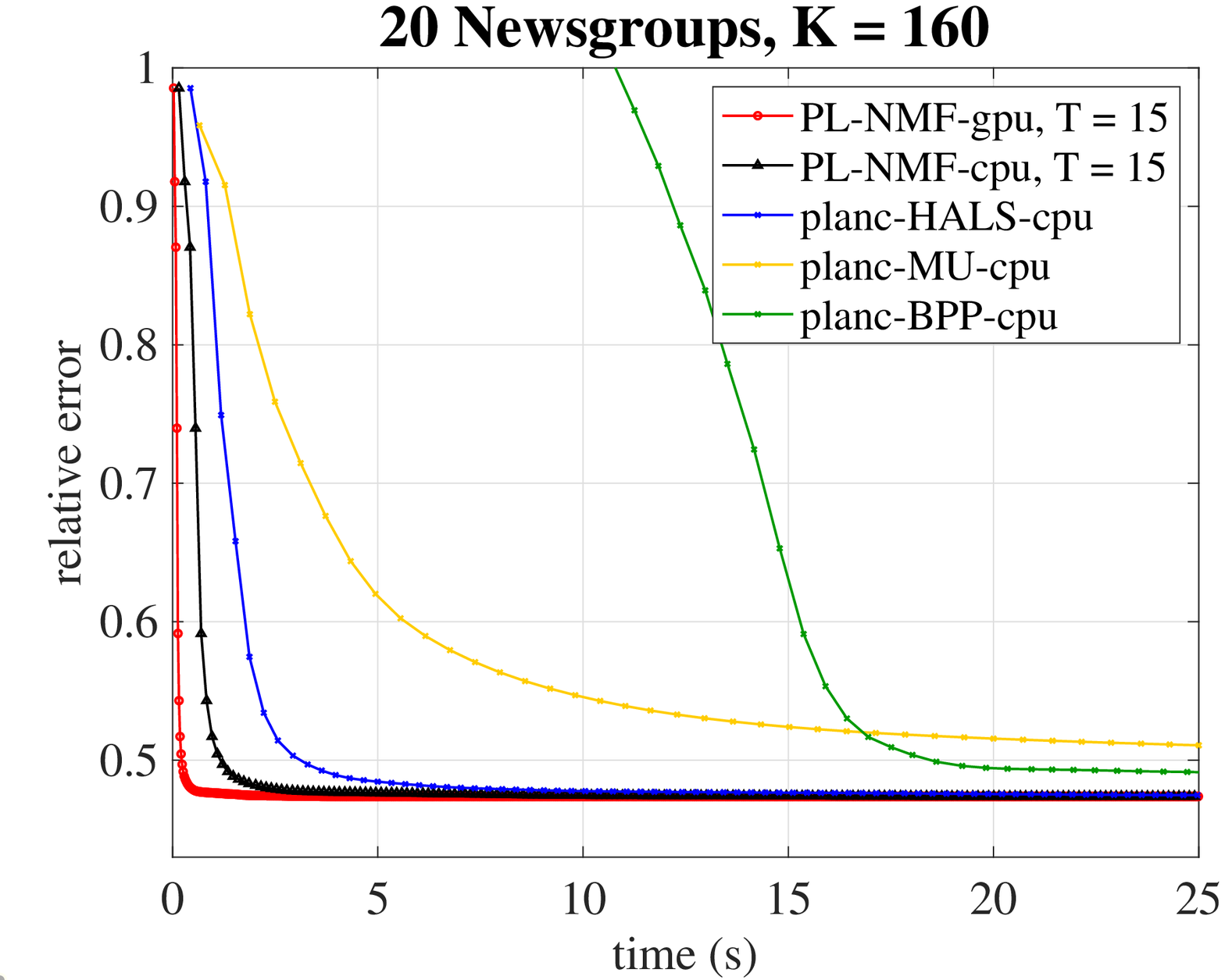}
  \includegraphics[width=0.3\linewidth]{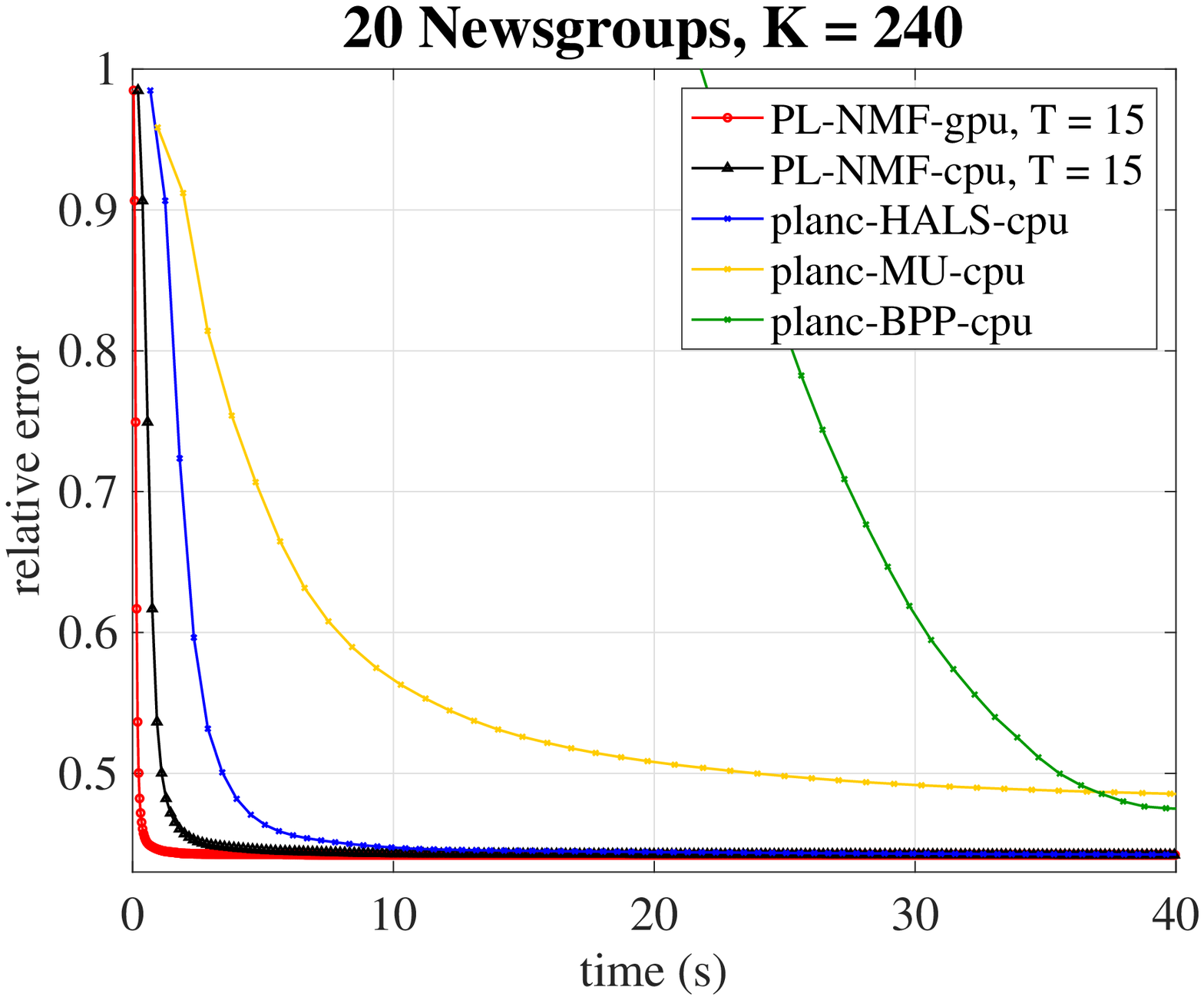}
  \includegraphics[width=0.3\linewidth]{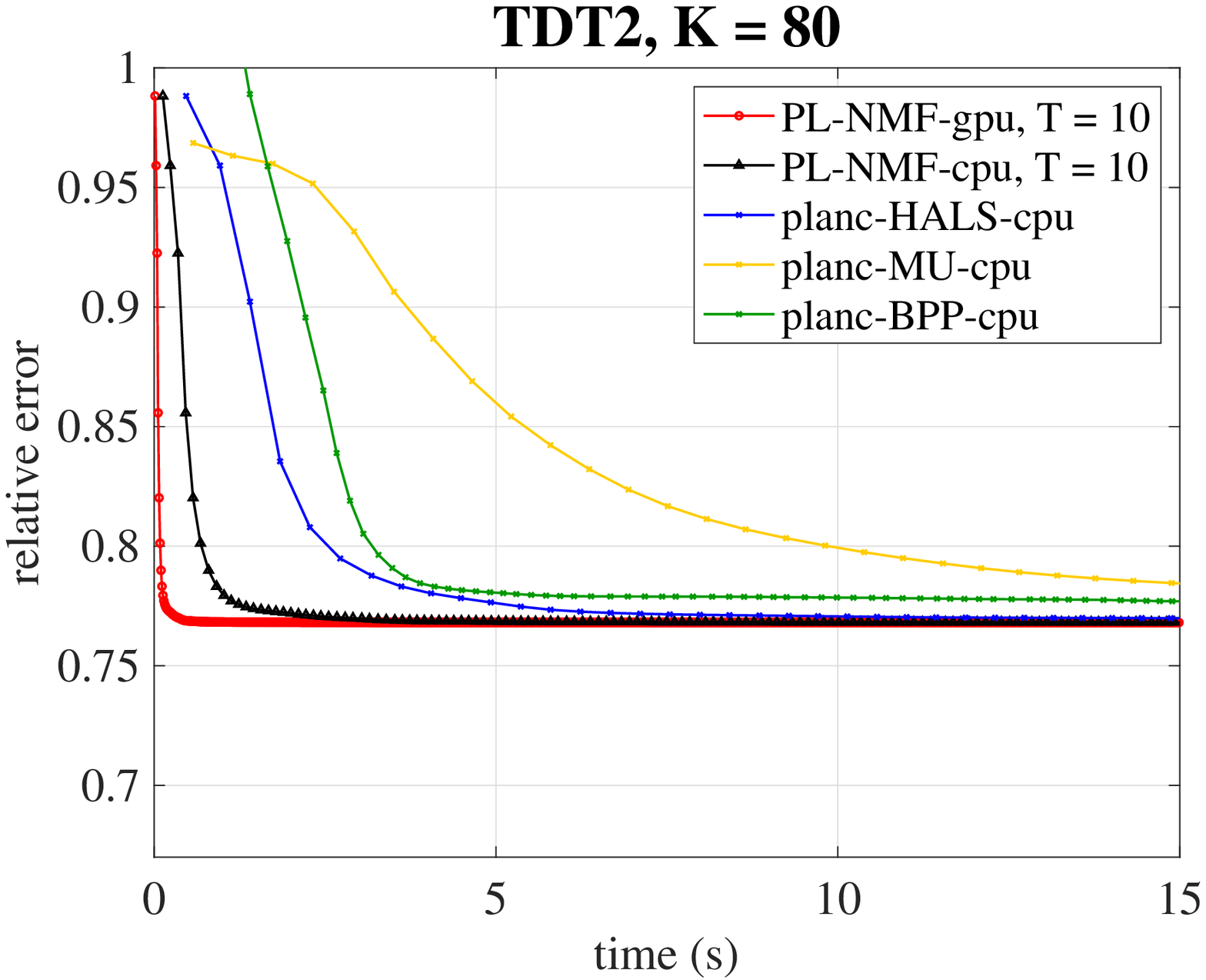}
  \includegraphics[width=0.3\linewidth]{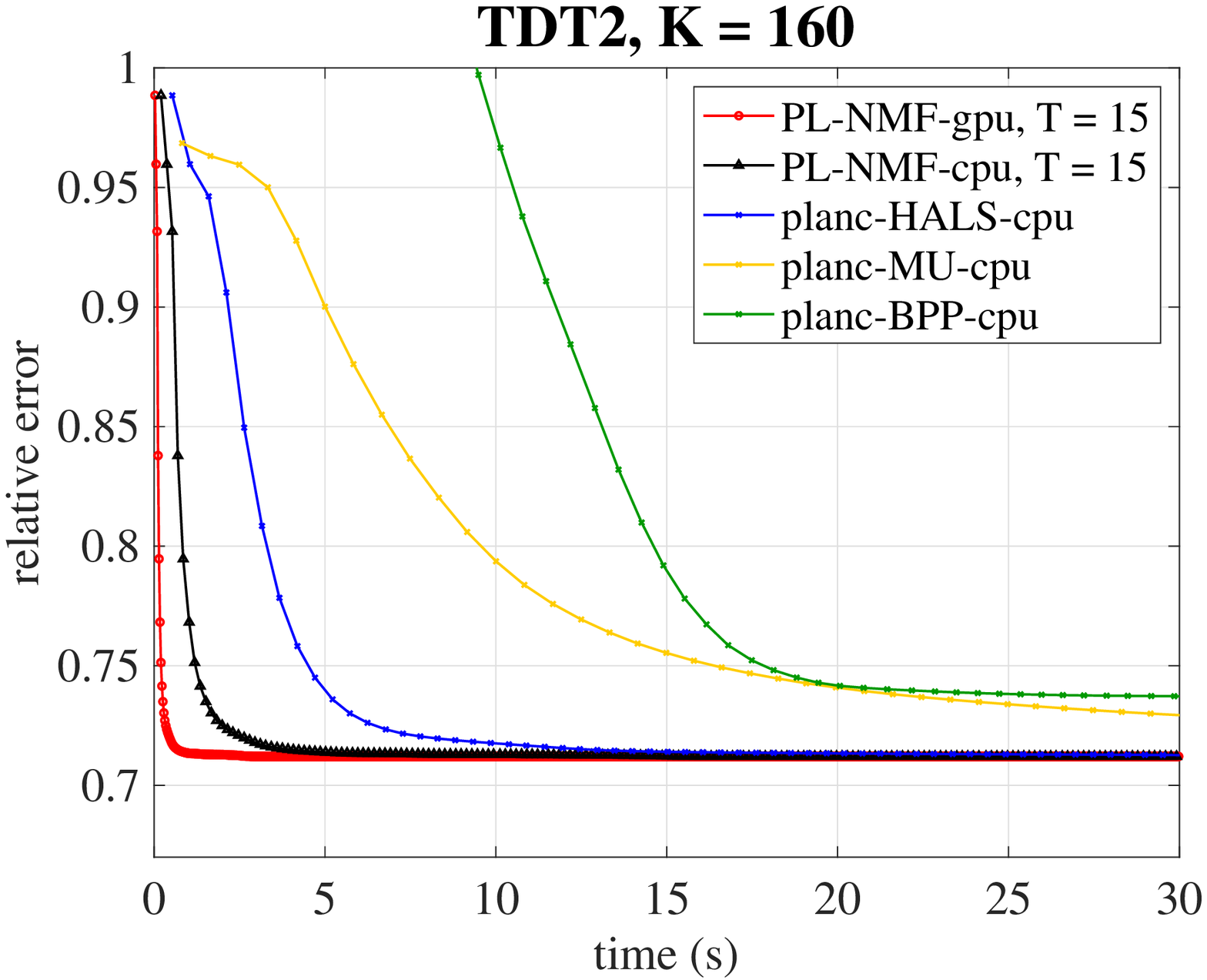}
  \includegraphics[width=0.3\linewidth]{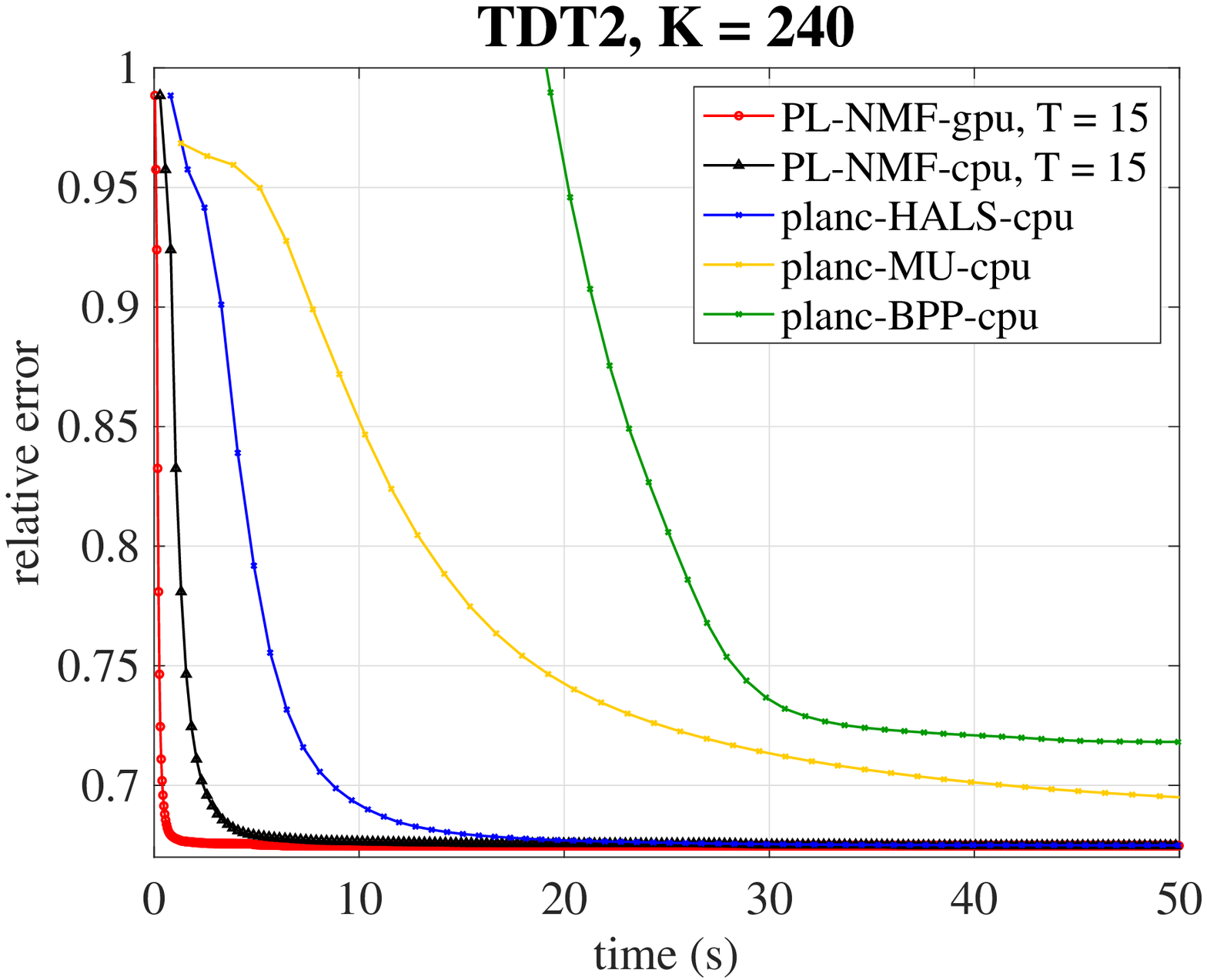}
  \includegraphics[width=0.3\linewidth]{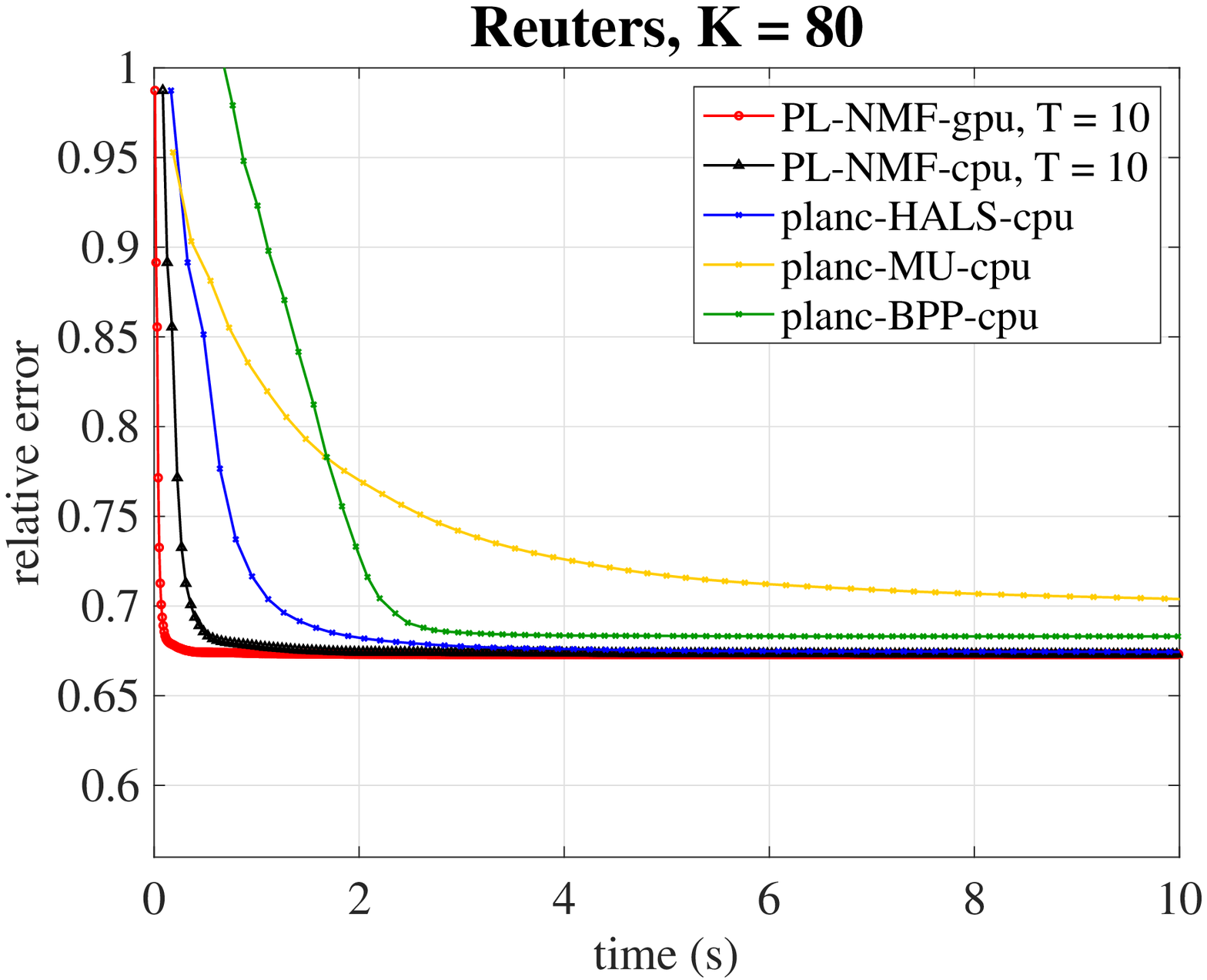}
  \includegraphics[width=0.3\linewidth]{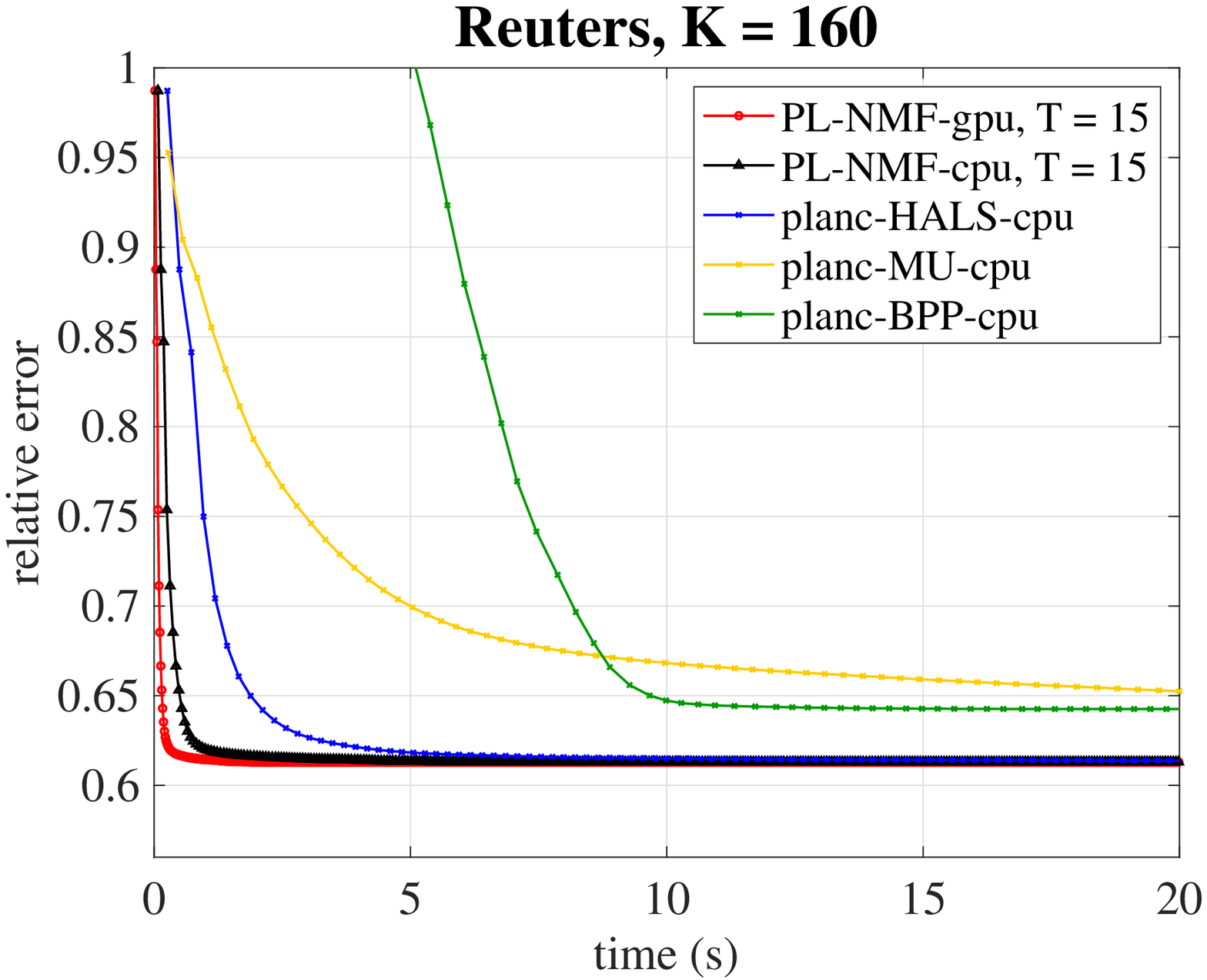}
  \includegraphics[width=0.3\linewidth]{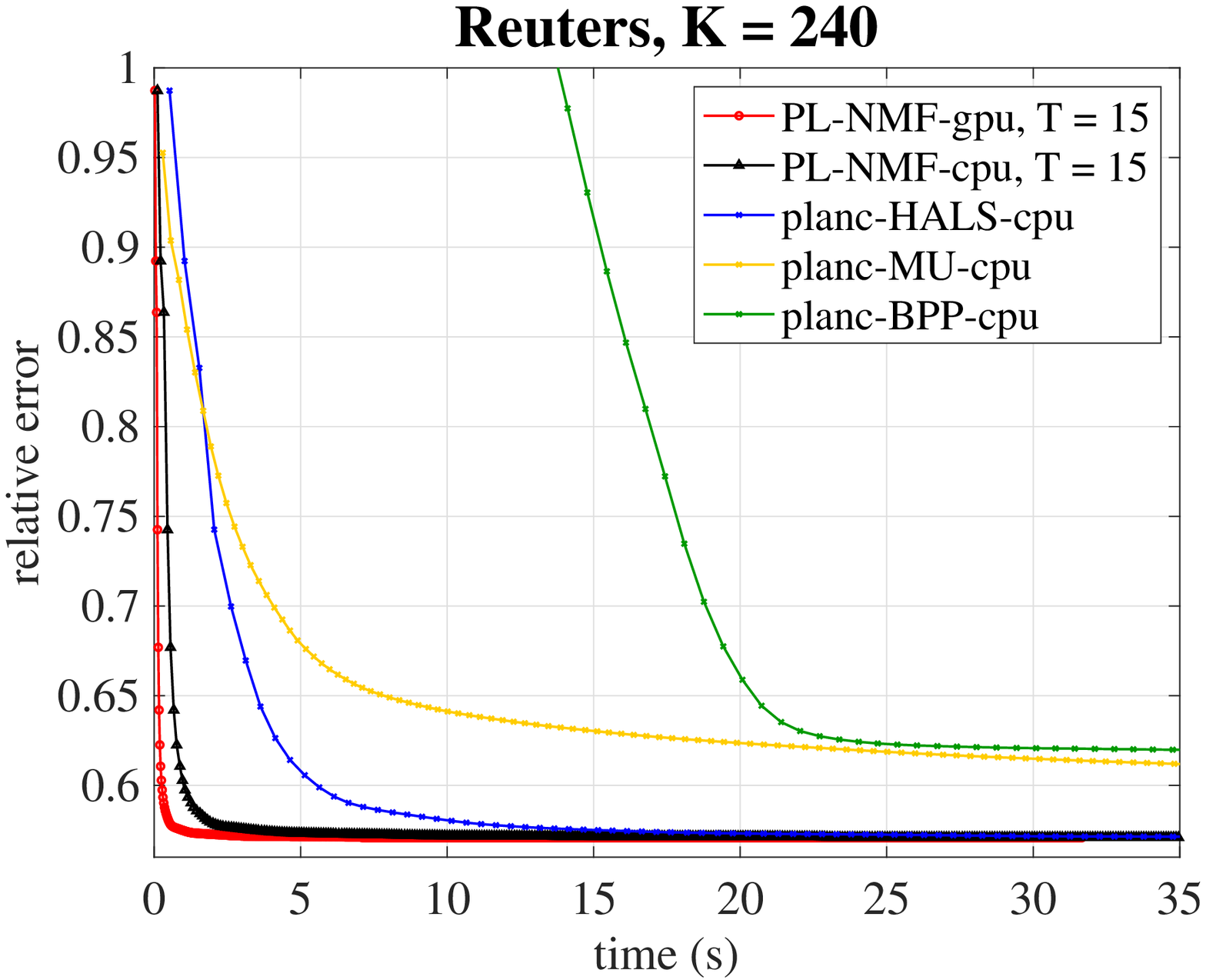}
  \includegraphics[width=0.3\linewidth]{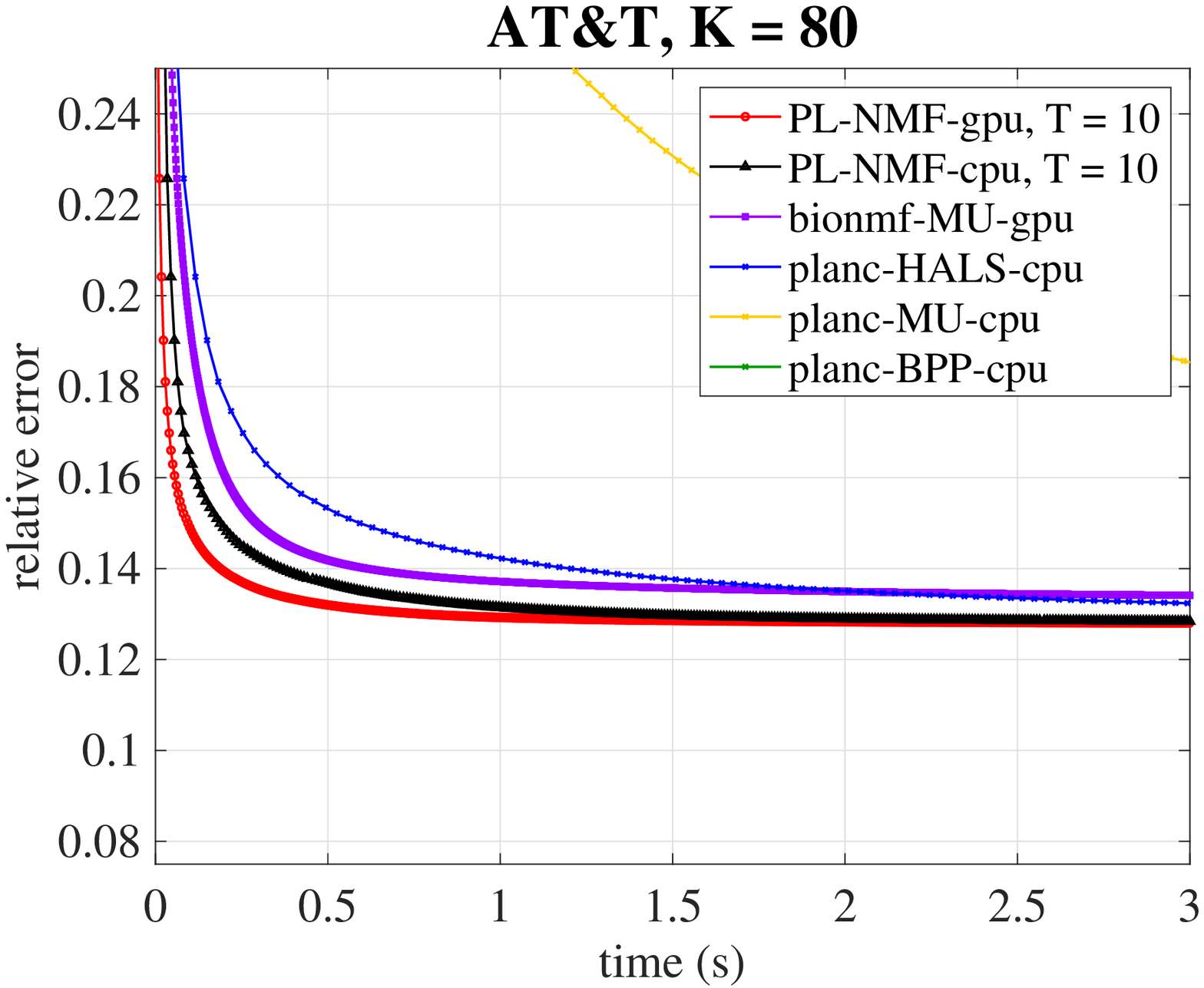}
  \includegraphics[width=0.3\linewidth]{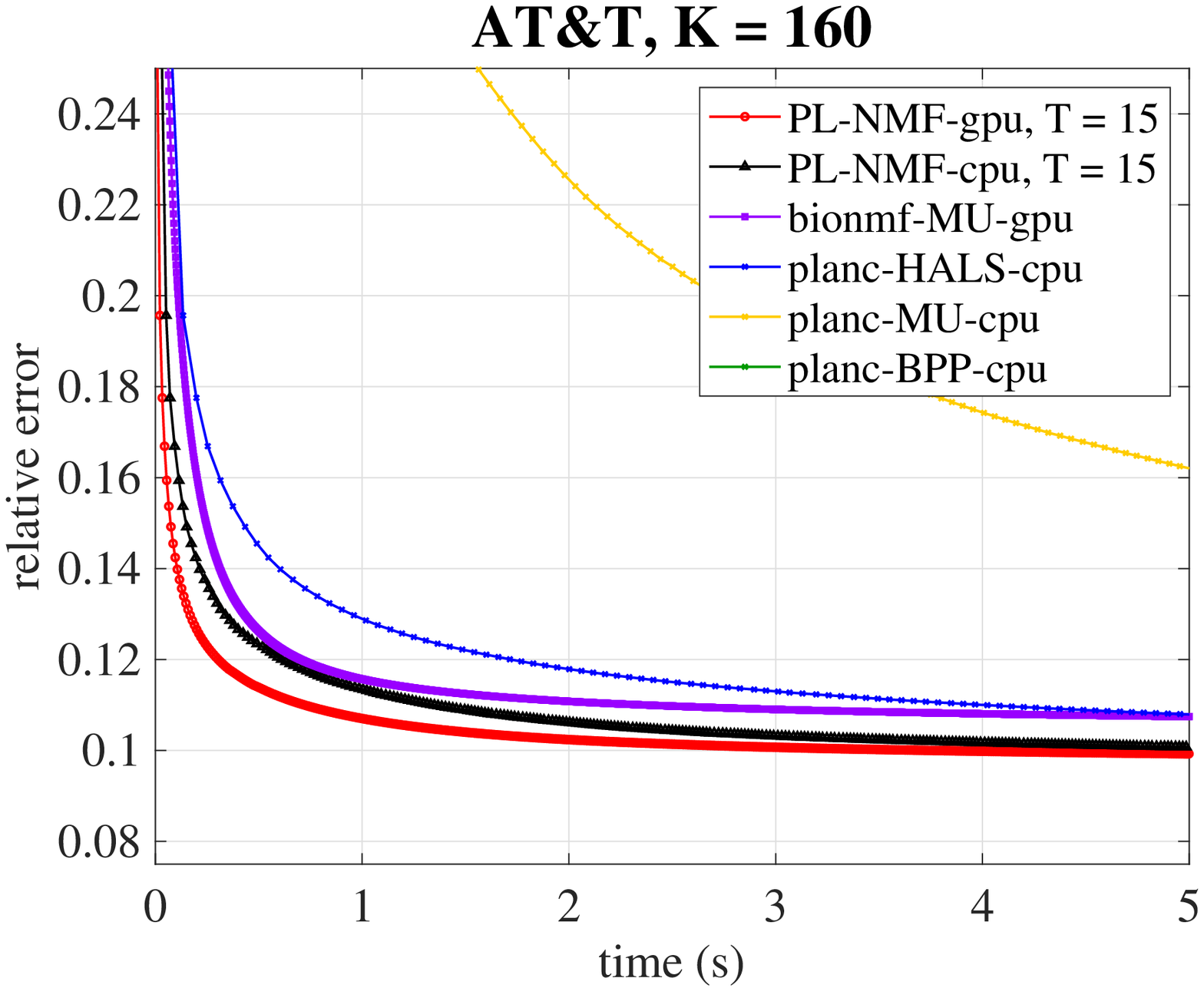}
  \includegraphics[width=0.3\linewidth]{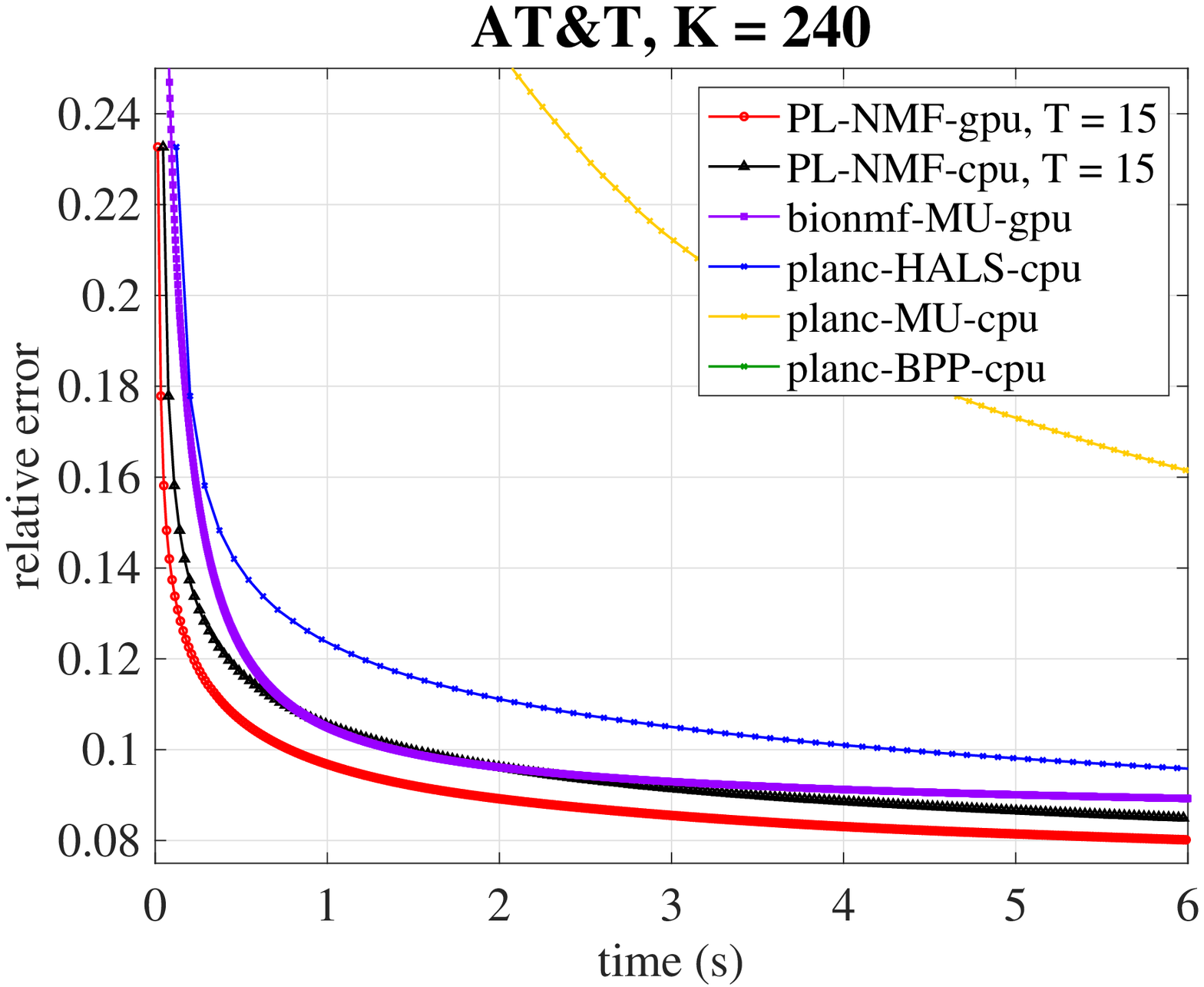}
  \includegraphics[width=0.3\linewidth]{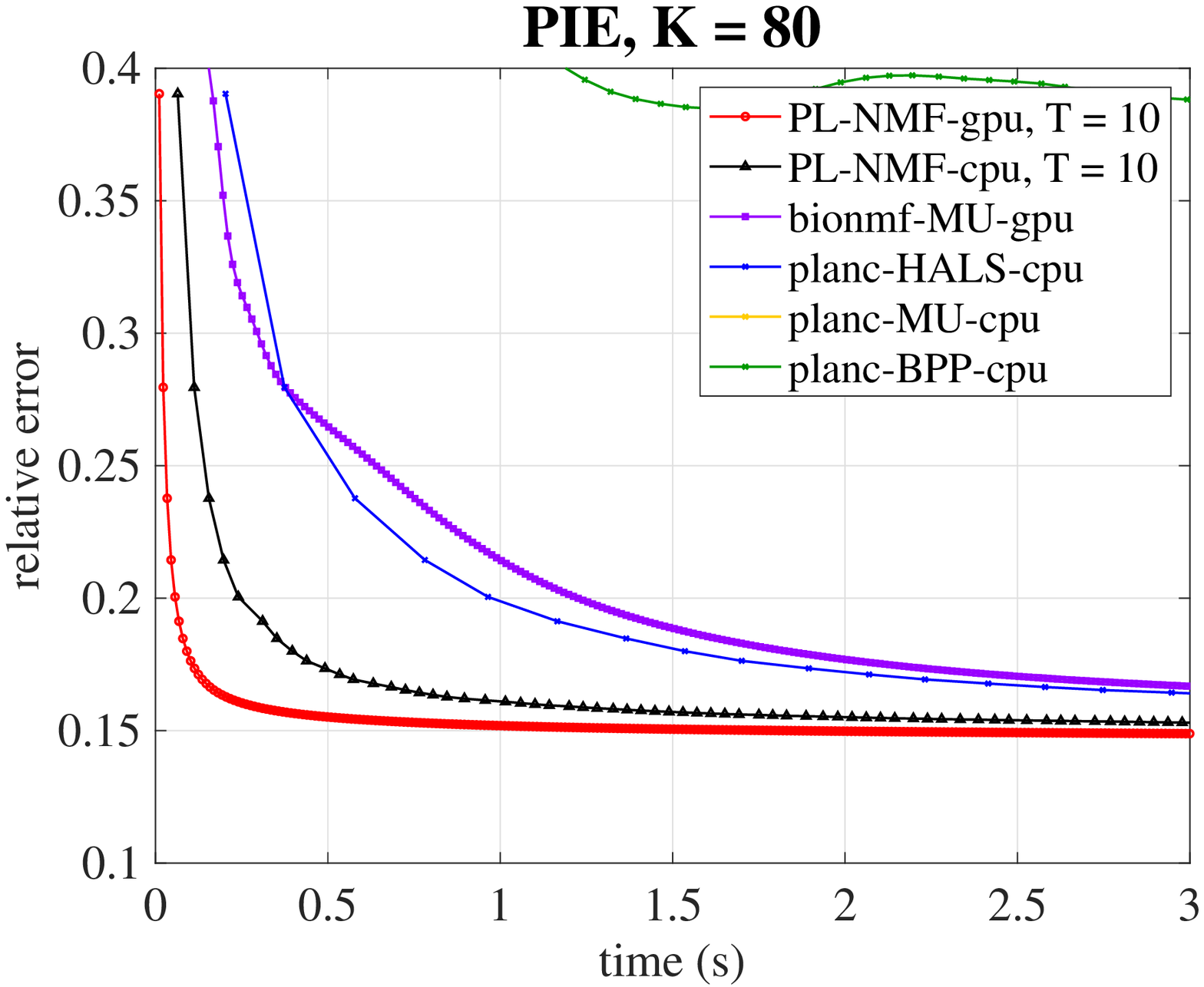}
  \includegraphics[width=0.3\linewidth]{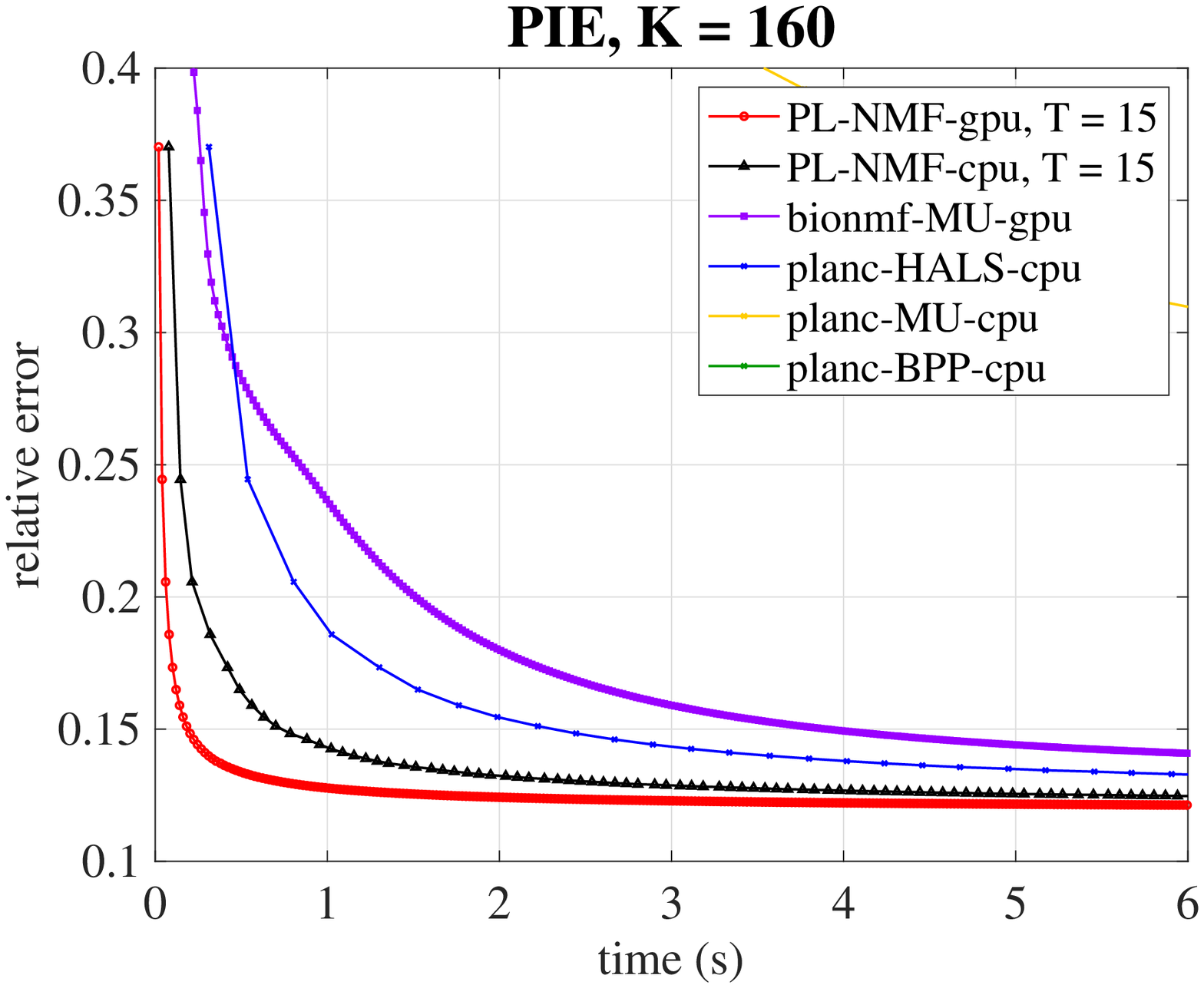}
  \includegraphics[width=0.3\linewidth]{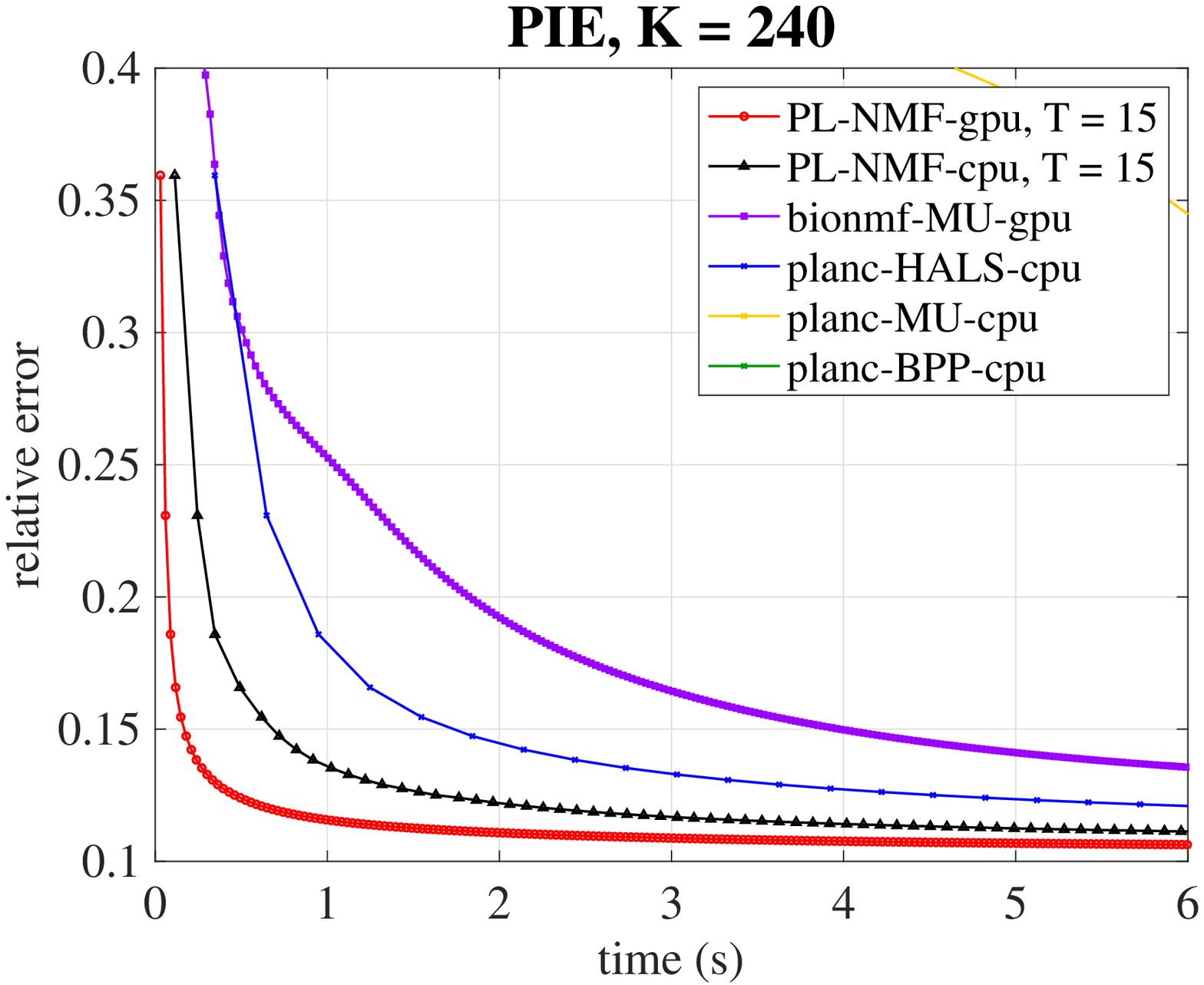}
  \caption{Relative objective value over time on five datasets. According to current model, the $T$ values for $K=$ 80, 160 and 240 are set to 10, 15 and 15, respectively. X-axis: elapsed time in seconds; Y-axis: relative error.}~\label{fig:nmf_con_time}
\end{figure*}
\begin{figure*}[h!]
\centering
  \includegraphics[width=0.21\linewidth]{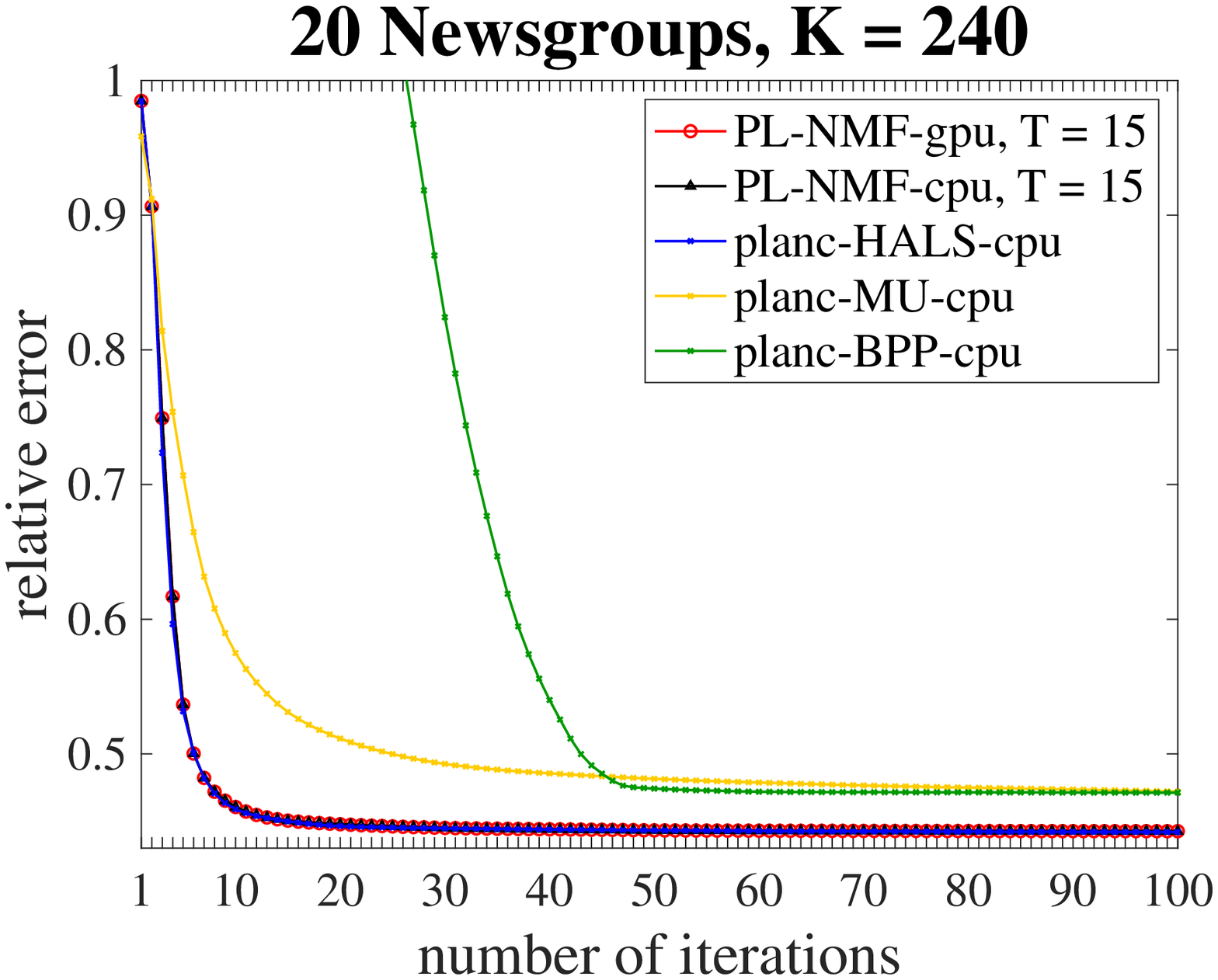}
  \hspace{-0.4cm}
  \includegraphics[width=0.21\linewidth]{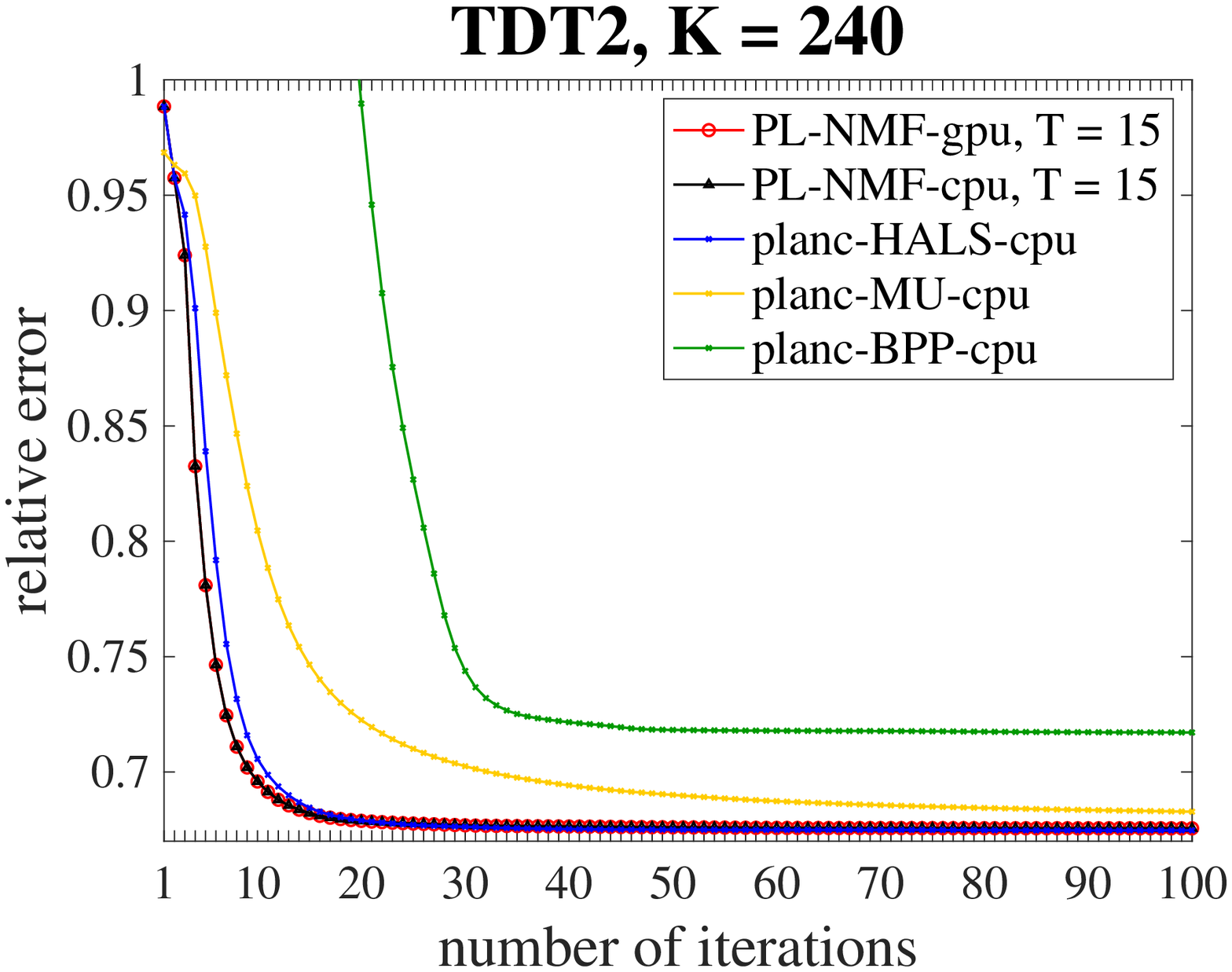}
  \hspace{-0.4cm}
  \includegraphics[width=0.21\linewidth]{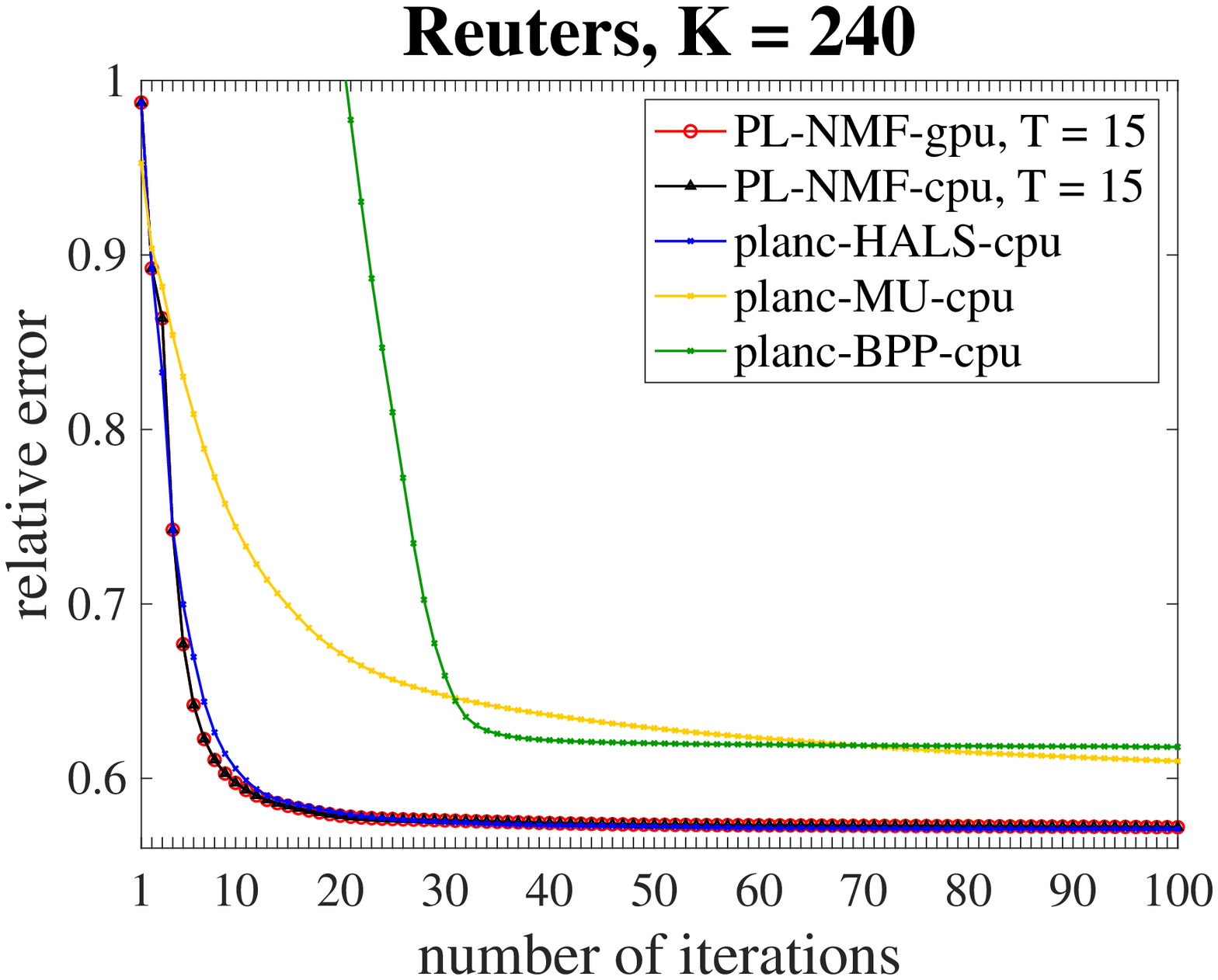}
  \hspace{-0.4cm}
  \includegraphics[width=0.21\linewidth]{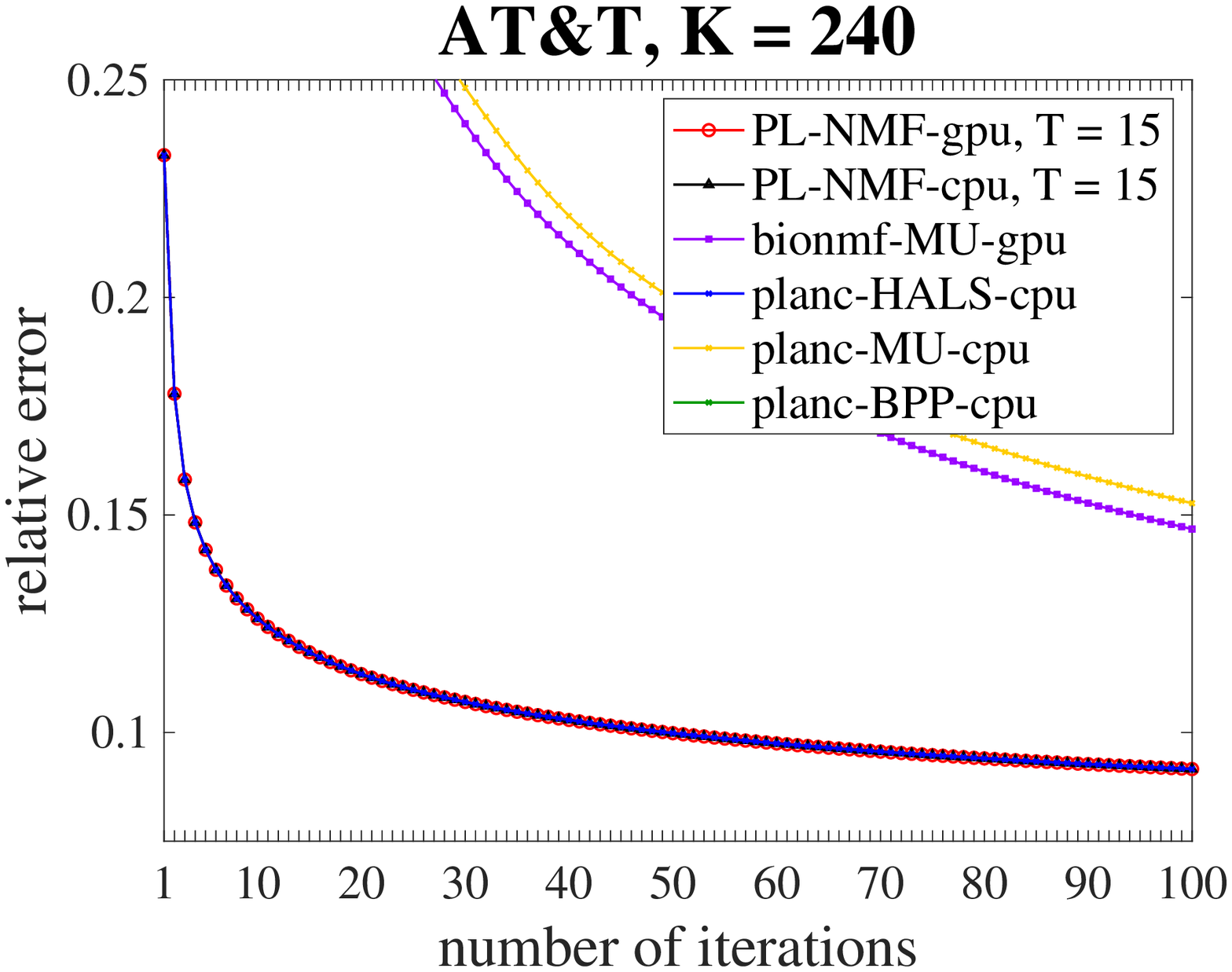}
  \hspace{-0.4cm}
  \includegraphics[width=0.21\linewidth]{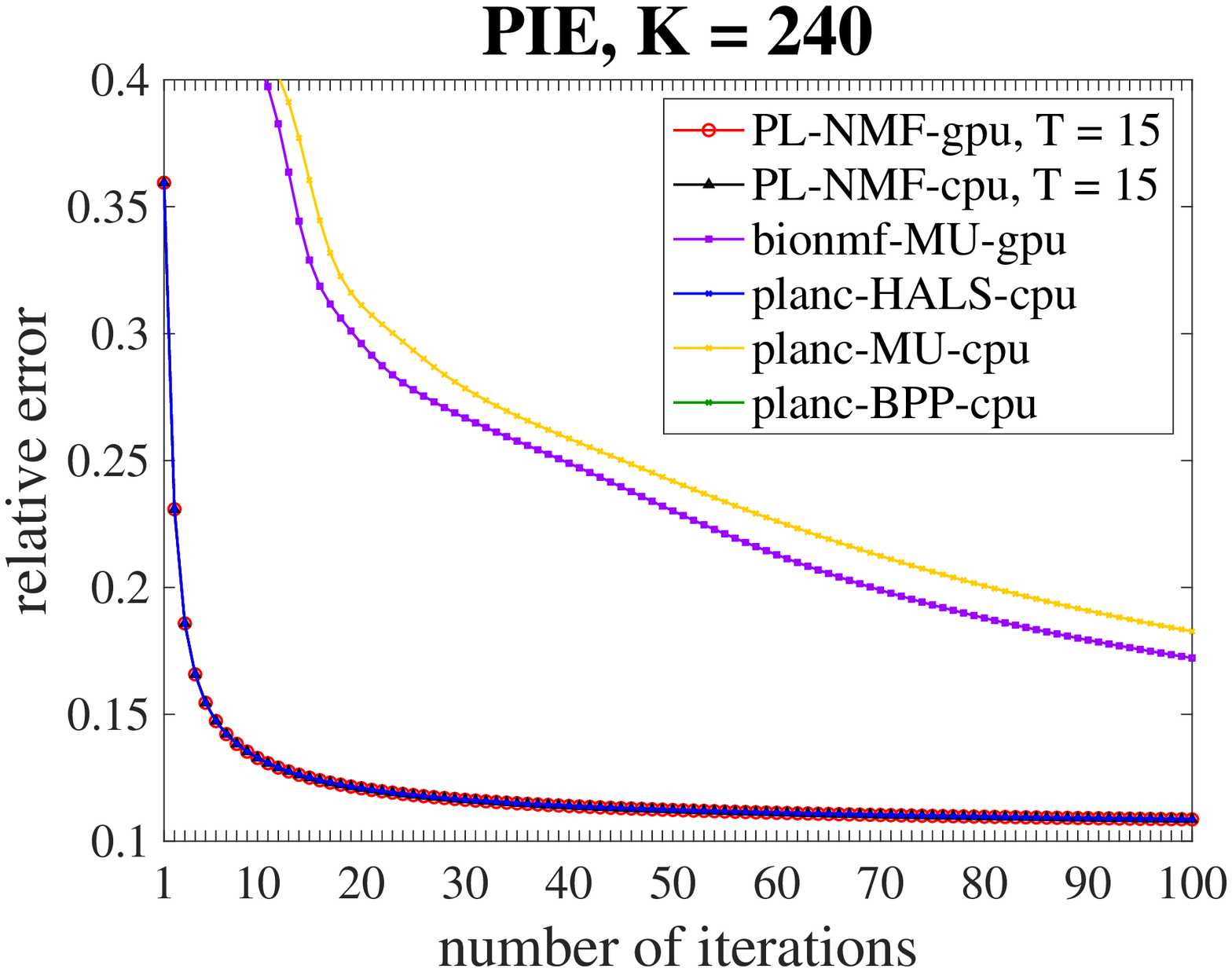}
  \caption{Comparison of convergence over iterations on five datasets, $K=$ 240 and $T=$ 15. X-axis: number of iterations; Y-axis: relative error.}~\label{fig:nmf_con_iter}
\end{figure*}
\begin{figure*}[h!]
\centering
  \includegraphics[width=0.21\linewidth]{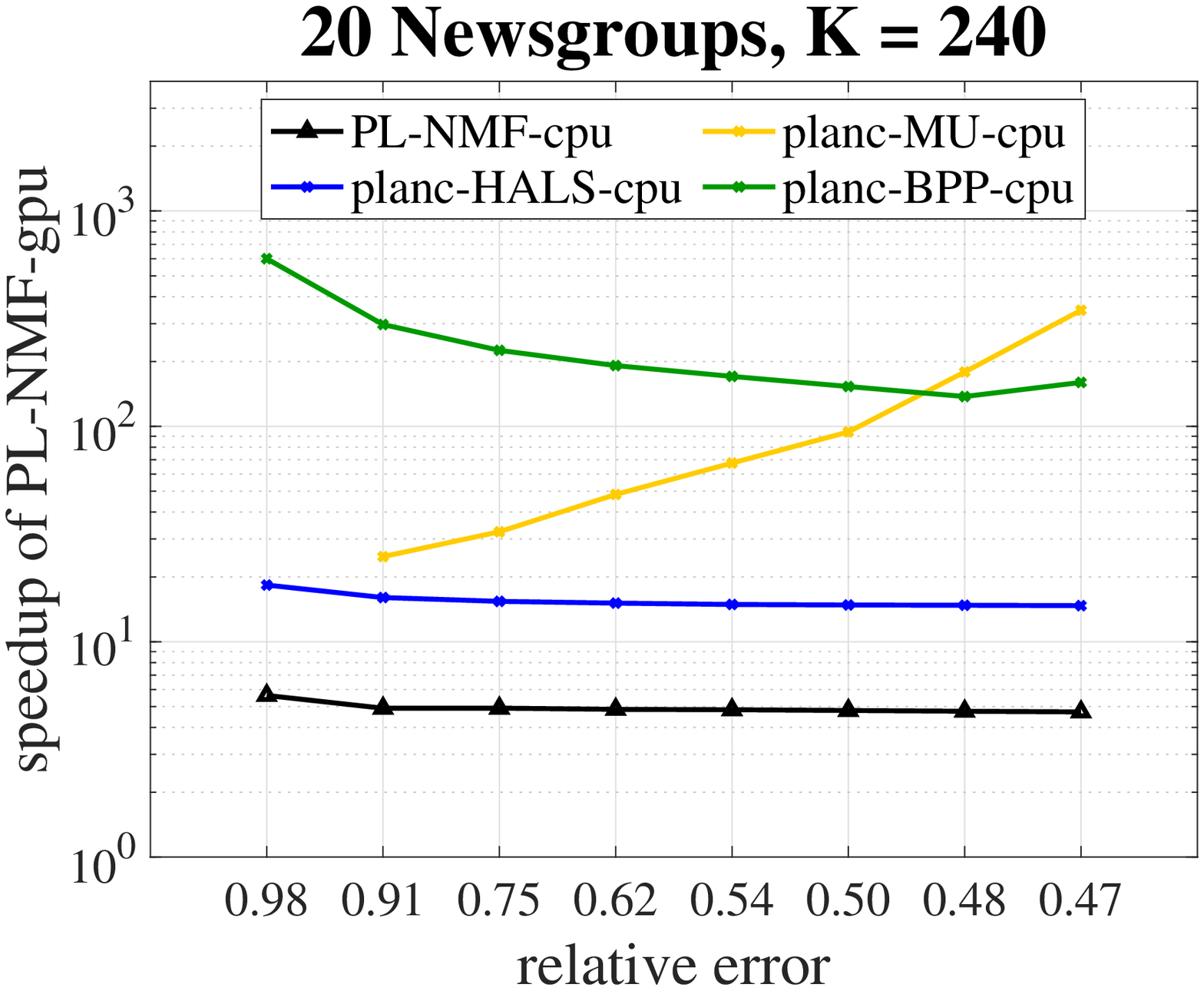}
   \hspace{-0.4cm}
  \includegraphics[width=0.21\linewidth]{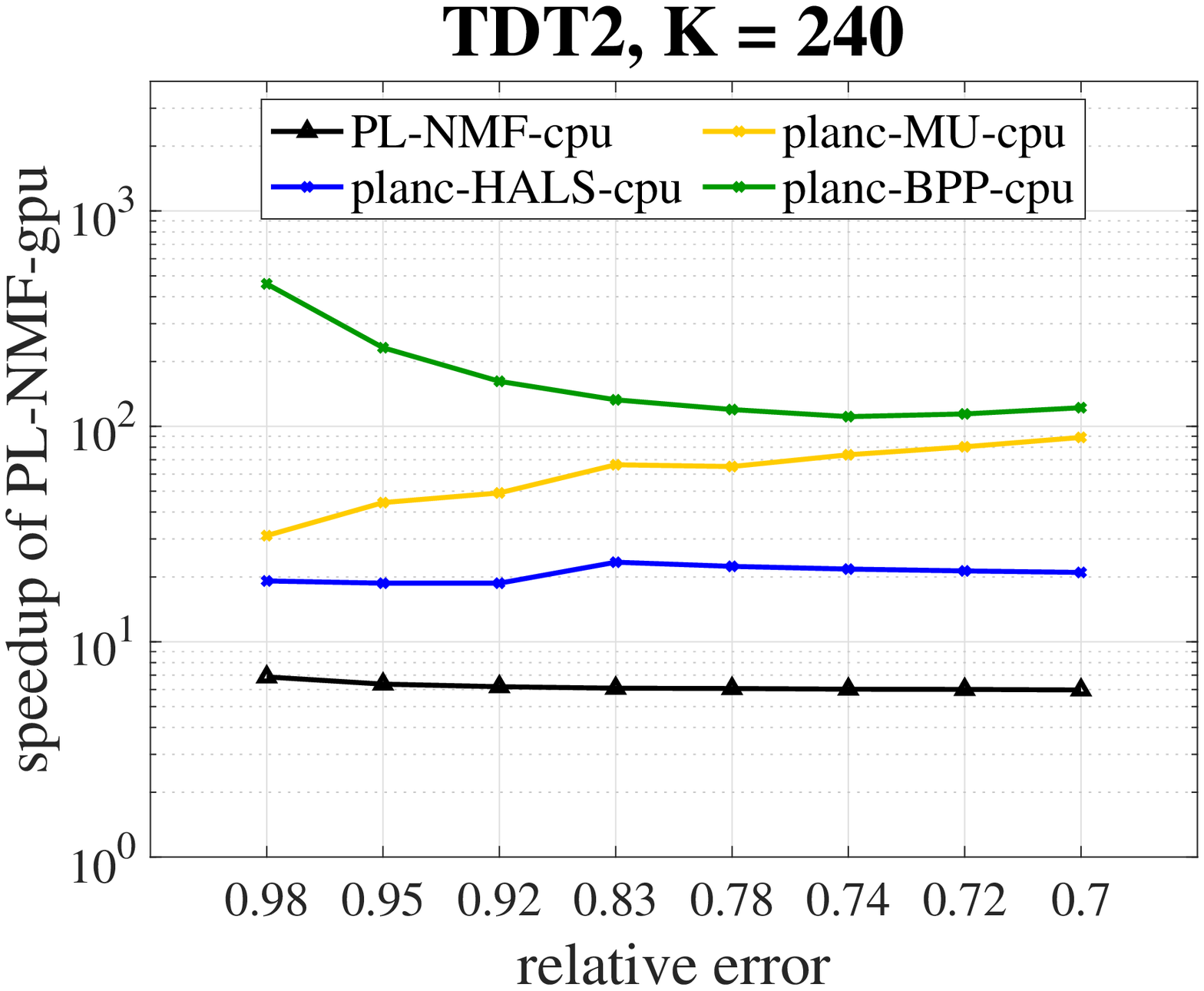}
   \hspace{-0.4cm}
  \includegraphics[width=0.21\linewidth]{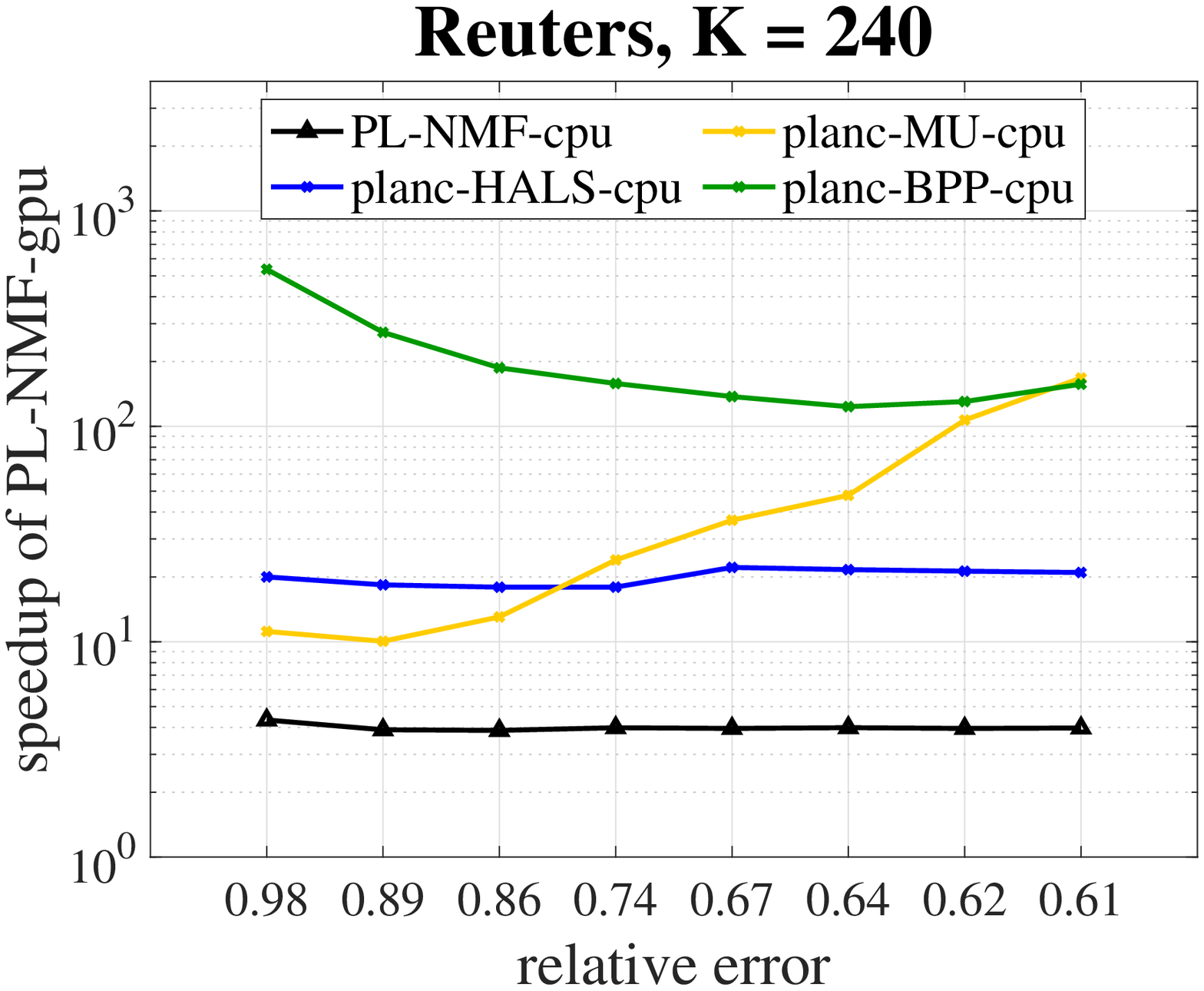}
   \hspace{-0.4cm}
 \includegraphics[width=0.21\linewidth]{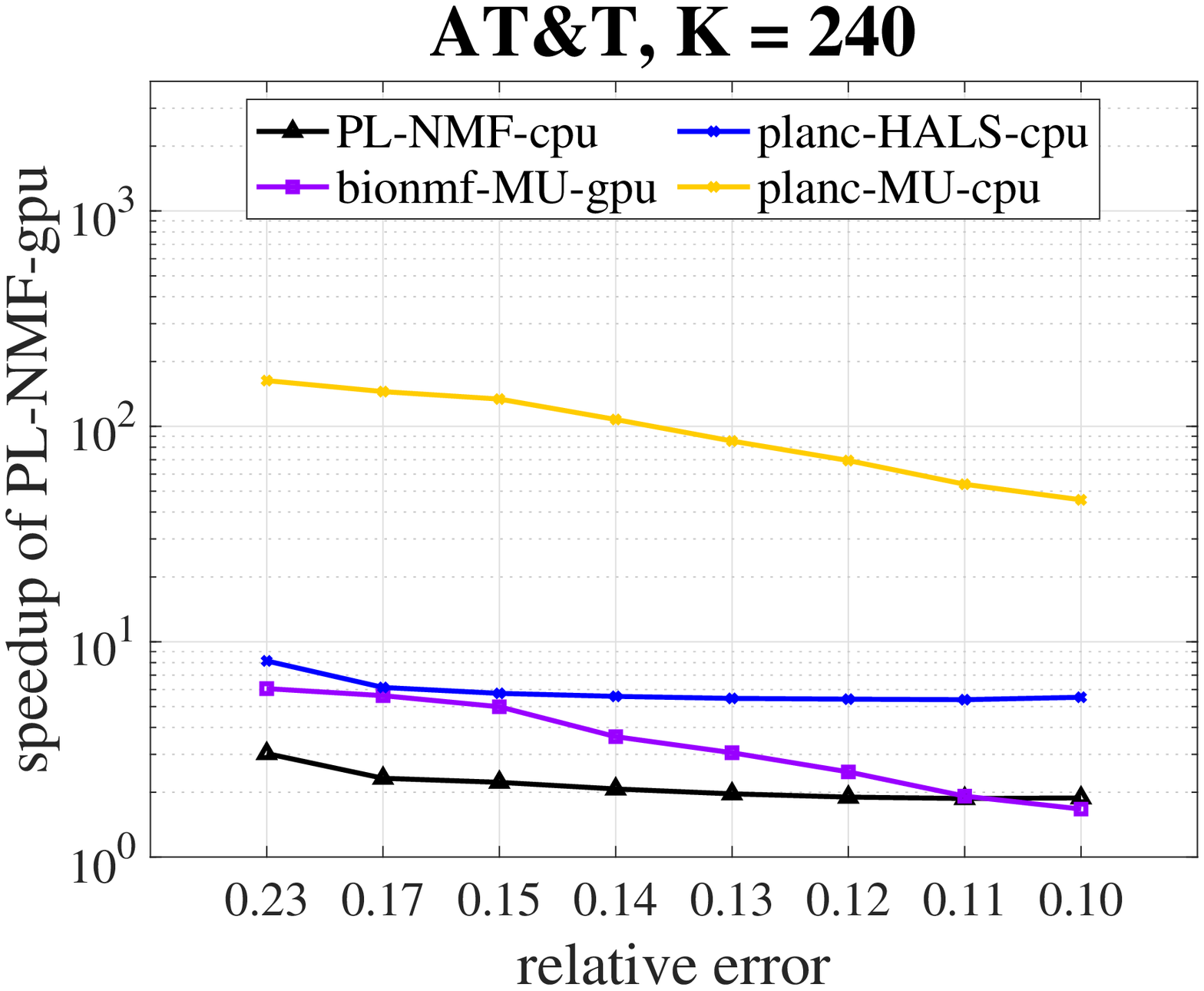}
  \hspace{-0.4cm}
  \includegraphics[width=0.21\linewidth]{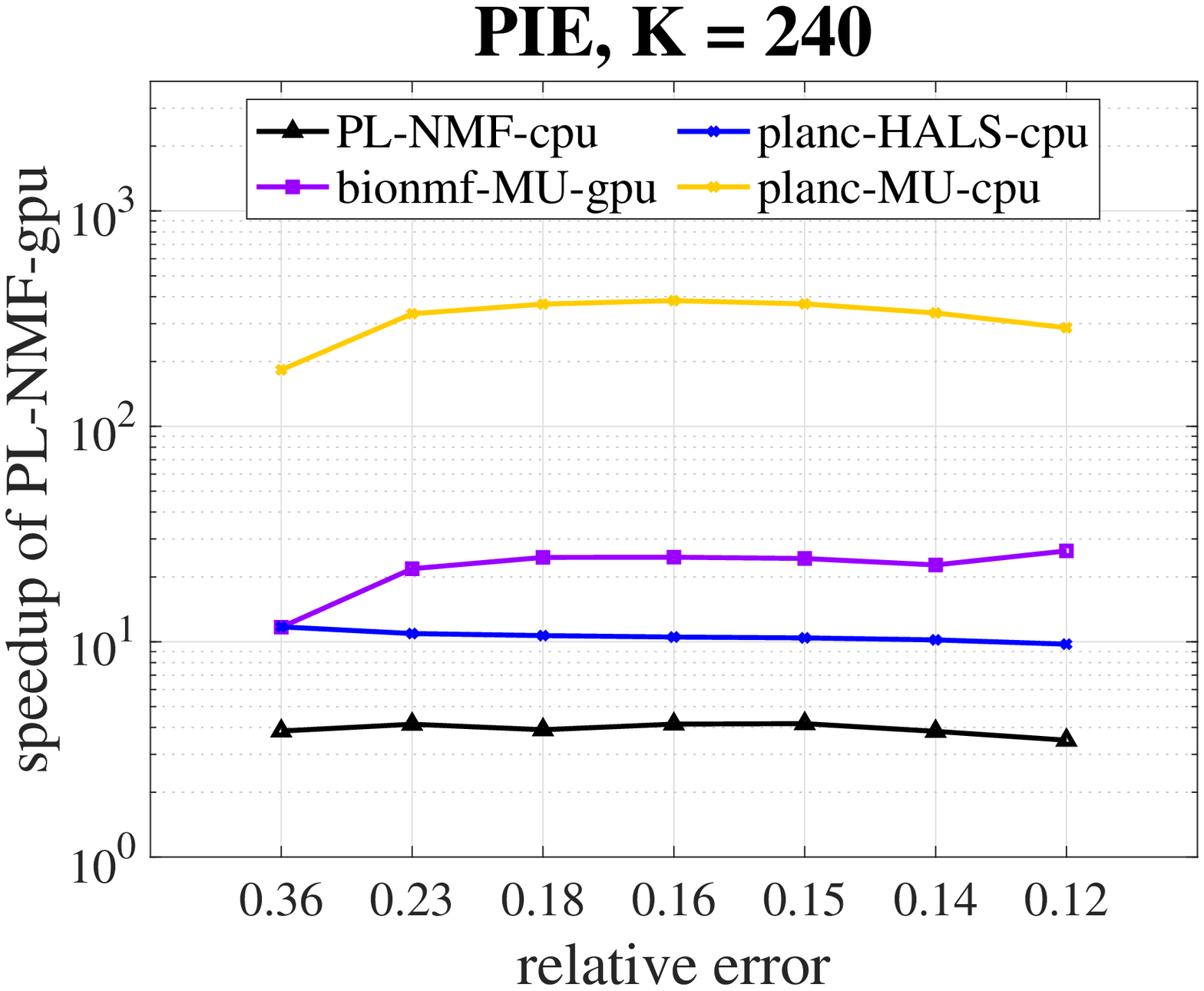}
  \caption{Speedup of PL-NMF-gpu over all CPU implementations on five datasets, $K=$ 240 and $T=$ 15.}~\label{fig:nmf_speedup}
\end{figure*}
\subsubsection{NMF Implementations Compared}
We evaluated PL-NMF on CPUs and GPUs with the state-of-the art parallel NMF
implementations such as planc\footnote{\scriptsize{https://github.com/ramkikannan/planc}} by
Kannan et al. \cite{kannan2016high,fairbanks2015behavioral} and
bionmf-gpu\footnote{\scriptsize{https://github.com/bioinfo-cnb/bionmf-gpu}} by Mej{\'\i}a-Roa
et al. \cite{mejia2015nmf}. The four implementations used in our comparisons are
as follows:

\setlist{nolistsep}
\begin{itemize}[noitemsep]
\item \textbf{planc-MU-cpu}: planc's OpenMP-based MU
\item \textbf{planc-HALS-cpu}: planc's OpenMP-based HALS
\item \textbf{planc-BPP-cpu}: planc's OpenMP-based ANLS-BPP
\item \textbf{bionmf-MU-gpu}: bionmf-gpu's GPU-based MU
\end{itemize}

All of the competing CPU implementations, including planc-MU-cpu, planc-HALS-cpu
and planc-BPP-cpu, and our PL-NMF-cpu, used Intel's Math Kernel Library (MKL) for
all BLAS (Basic Linear Algebra Subprograms) operations. Similarly, all GPU
implementations, including bionmf-MU-gpu and our PL-NMF-gpu, used NVIDIA's cuBLAS
library for all types of BLAS operations.

\subsubsection{Evaluation Metric}
\hfill\\
In order to evaluate the accuracy of different NMF models, we used the relative
objective function  $\sqrt{\frac{\sum_{vd} (A_{vd}-(WH)_{vd})^{2}}{\sum_{vd}
(A_{vd})^{2}}}$ suggested by Kim et al. \cite{kim2011fast}, where $A_{vd}$ and $(WH)_{vd}$ denote the values of each element in an input matrix $A \in \mathbb R_+^{V \times D}$ and an approximated matrix $(WH) \in \mathbb R_+^{V \times D}$. The capability of each NMF model in minimizing the objective function can be obtained by measuring relative changes of objective value over iterations.

\subsection{Performance Evaluation}
\subsubsection{Convergence}
\hfill\\
Figure \ref{fig:nmf_con_time} shows the relative error as a function of elapsed
time of various NMF implementations for different $K$ values. To ensure
fairness, the number of threads in all CPU implementations were tuned per
dataset and the best performing configuration was selected. For each dataset,
the same randomly initialized non-negative matrices were used for all CPU and
GPU implementations. Since the bionmf-MU-gpu implementation does not allow the input
matrix to be sparse, we only compared our GPU implementation with bionmf-MU-gpu
on AT\&T and PIE dense image datasets. PL-NMF-cpu and PL-NMF-gpu consistently
outperformed existing state-of-the-art CPU and GPU implementations on all
datasets. As reported in previous studies, FAST-HALS produced a better convergence
rate than other NMF variants. MU and ANLS-BPP algorithms suffered from a lower
convergence rate on both sparse and dense matrices. As shown in Figure
\ref{fig:nmf_con_iter}, planc-HALS-cpu was the only implementation which was 
able to maintain the same solution quality as ours. However, our implementation
converged faster. 

\subsubsection{Speedup}
\hfill\\
Compared to the planc-HALS-cpu, our PL-NMF-cpu achieved 3.07$\times$,
3.06$\times$, 5.81$\times$, 3.02$\times$ and 3.07$\times$ speedup per iteration
on the 20 Newsgroups, TDT2, Reuters, AT\&T and PIE datasets with $K$ = 240,
respectively.  As the relative error reduction per iteration is vastly different
between MU and FAST-HALS algorithms, measuring the speedup per iteration
between bionmf-MU-gpu and PL-NMF-gpu is not a fair comparison. 

Figure \ref{fig:nmf_speedup} depicts the speedup of our PL-NMF-gpu over all CPU
implementations. The x-axis in Figure \ref{fig:nmf_speedup} is relative error,
and the y-axis is the ratio of elapsed time for all CPU implementations to reach
a relative error to elapsed time for PL-NMF-gpu to approach the same relative
error. All of the points in Figure \ref{fig:nmf_speedup} are greater than one.
This indicates that PL-NMF-gpu is faster than all of the competing
implementations. For example, when the compared models, i.e.,
PL-NMF-cpu, planc-HALS-cpu, bionmf-MU-gpu and planc-MU-cpu,
converged to 0.12 relative error, the parallel PL-NMF-gpu achieved 3.49$\times$,
9.74$\times$, 26.41$\times$ and 287.1$\times$ speedup on PIE dataset,
respectively.

\begin{table}[h]
\centering
\caption{Breakdown of elapsed time in seconds for updating $W$ on the 20 Newsgroups dataset. DMV: Iterative Dense
		Matrix-Vector Multiplications; DMM: Dense Matrix-Dense Matrix Multiplication;
		SpMM: Sparse Matrix-Dense Matrix Multiplication.}
\label{tb:breakdown}
\scalebox{0.8}{
\begin{tabular}{|c|c|c|c|}
\hline
\begin{tabular}[c]{@{}c@{}}\textbf{Sequential} \\ \textbf{FAST-HALS
NMF}\end{tabular} & \textbf{elapsed time (s)} & \textbf{PL-NMF-cpu}  &
\textbf{elapsed time (s)} \\ \hline
SpMM                                   & 0.048                      & SpMM             & 0.048     \\ \hline
DMM                                    & 0.002                      & DMM              & 0.002     \\ \hline
\multirow{2}{*}{DMV}                   & \multirow{2}{*}{2.039}     & Phase 1          & 0.005     \\ \cline{3-4}
                                       &                            & Phase 2 \& 3     & 0.026     \\ \hline
\end{tabular}}
\end{table}

Table \ref{tb:breakdown} shows the breakdown of
elapsed time for each step in updating $W$. Both sequential FAST-HALS NMF and
PL-NMF-cpu implementations use the same mkl\_dcsrmm() and cblas\_dgemm() routines
for SpMM and DMM operations. In Table \ref{tb:breakdown}, SpMM corresponds to
line 10 in Algorithm 1 and line 1 in Algorithm 2, which computes the same
$AH^{T}$. Similarly, DMM corresponds to line 11 in Algorithm 1 and line 2 in
Algorithm 2, which performs the same $HH^{T}$. The difference of updating $W$ is
that PL-NMF-cpu performs phases 1, 2 and 3 instead of iteratively performing DMV
computations. As expected, the updating time of $W$ is considerably decreased in our
PL-NMF-cpu algorithm, indicating that the reformulation of the core-computations to
matrix-matrix multiplication shows the benefit of our approach.

\section{Conclusion}
\label{section:conclusion}

In this paper, we developed a HALS-based parallel NMF algorithm for multi-core
CPUs and GPUs. The data movement overhead is a critical factor that affects
performance. This paper does a systematic analysis of data movement overheads
associated with NMF algorithm to determine the bottlenecks. Our proposed
approach alleviates the data movement overheads by enhancing data locality. Our
experimental section shows that our parallel NMF achieves significant
performance improvement over the existing state-of-the-art parallel
implementations.

%
\bibliographystyle{ACM-Reference-Format}
\bibliography{references}


\begin{thebibliography}{27}


\ifx \showCODEN    \undefined \def \showCODEN     #1{\unskip}     \fi
\ifx \showDOI      \undefined \def \showDOI       #1{#1}\fi
\ifx \showISBNx    \undefined \def \showISBNx     #1{\unskip}     \fi
\ifx \showISBNxiii \undefined \def \showISBNxiii  #1{\unskip}     \fi
\ifx \showISSN     \undefined \def \showISSN      #1{\unskip}     \fi
\ifx \showLCCN     \undefined \def \showLCCN      #1{\unskip}     \fi
\ifx \shownote     \undefined \def \shownote      #1{#1}          \fi
\ifx \showarticletitle \undefined \def \showarticletitle #1{#1}   \fi
\ifx \showURL      \undefined \def \showURL       {\relax}        \fi
\providecommand\bibfield[2]{#2}
\providecommand\bibinfo[2]{#2}
\providecommand\natexlab[1]{#1}
\providecommand\showeprint[2][]{arXiv:#2}

\bibitem[\protect\citeauthoryear{Aghdam, Analoui, and Kabiri}{Aghdam
  et~al\mbox{.}}{2015}]%
        {aghdam2015novel}
\bibfield{author}{\bibinfo{person}{Mehdi~Hosseinzadeh Aghdam},
  \bibinfo{person}{Morteza Analoui}, {and} \bibinfo{person}{Peyman Kabiri}.}
  \bibinfo{year}{2015}\natexlab{}.
\newblock \showarticletitle{A novel non-negative matrix factorization method
  for recommender systems}.
\newblock \bibinfo{journal}{\emph{Applied Mathematics \& Information Sciences}}
  \bibinfo{volume}{9}, \bibinfo{number}{5} (\bibinfo{year}{2015}),
  \bibinfo{pages}{2721}.
\newblock


\bibitem[\protect\citeauthoryear{Battenberg and Wessel}{Battenberg and
  Wessel}{2009}]%
        {battenberg2009accelerating}
\bibfield{author}{\bibinfo{person}{Eric Battenberg} {and}
  \bibinfo{person}{David Wessel}.} \bibinfo{year}{2009}\natexlab{}.
\newblock \showarticletitle{Accelerating Non-Negative Matrix Factorization for
  Audio Source Separation on Multi-Core and Many-Core Architectures.}. In
  \bibinfo{booktitle}{\emph{ISMIR}}. \bibinfo{pages}{501--506}.
\newblock


\bibitem[\protect\citeauthoryear{Cichocki and Phan}{Cichocki and Phan}{2009}]%
        {cichocki2009fast}
\bibfield{author}{\bibinfo{person}{Andrzej Cichocki} {and}
  \bibinfo{person}{Anh-Huy Phan}.} \bibinfo{year}{2009}\natexlab{}.
\newblock \showarticletitle{Fast local algorithms for large scale nonnegative
  matrix and tensor factorizations}.
\newblock \bibinfo{journal}{\emph{IEICE transactions on fundamentals of
  electronics, communications and computer sciences}} \bibinfo{volume}{92},
  \bibinfo{number}{3} (\bibinfo{year}{2009}), \bibinfo{pages}{708--721}.
\newblock


\bibitem[\protect\citeauthoryear{Cichocki, Zdunek, and Amari}{Cichocki
  et~al\mbox{.}}{2007}]%
        {cichocki2007hierarchical}
\bibfield{author}{\bibinfo{person}{Andrzej Cichocki}, \bibinfo{person}{Rafal
  Zdunek}, {and} \bibinfo{person}{Shun-ichi Amari}.}
  \bibinfo{year}{2007}\natexlab{}.
\newblock \showarticletitle{Hierarchical ALS algorithms for nonnegative matrix
  and 3D tensor factorization}. In \bibinfo{booktitle}{\emph{International
  Conference on Independent Component Analysis and Signal Separation}}.
  Springer, \bibinfo{pages}{169--176}.
\newblock


\bibitem[\protect\citeauthoryear{Dong, Zhao, and Wang}{Dong
  et~al\mbox{.}}{2010}]%
        {dong2010parallel}
\bibfield{author}{\bibinfo{person}{Chao Dong}, \bibinfo{person}{Huijie Zhao},
  {and} \bibinfo{person}{Wei Wang}.} \bibinfo{year}{2010}\natexlab{}.
\newblock \showarticletitle{Parallel nonnegative matrix factorization algorithm
  on the distributed memory platform}.
\newblock \bibinfo{journal}{\emph{International journal of parallel
  programming}} \bibinfo{volume}{38}, \bibinfo{number}{2}
  (\bibinfo{year}{2010}), \bibinfo{pages}{117--137}.
\newblock


\bibitem[\protect\citeauthoryear{Fairbanks, Kannan, Park, and Bader}{Fairbanks
  et~al\mbox{.}}{2015}]%
        {fairbanks2015behavioral}
\bibfield{author}{\bibinfo{person}{James~P Fairbanks},
  \bibinfo{person}{Ramakrishnan Kannan}, \bibinfo{person}{Haesun Park}, {and}
  \bibinfo{person}{David~A Bader}.} \bibinfo{year}{2015}\natexlab{}.
\newblock \showarticletitle{Behavioral clusters in dynamic graphs}.
\newblock \bibinfo{journal}{\emph{Parallel Comput.}}  \bibinfo{volume}{47}
  (\bibinfo{year}{2015}), \bibinfo{pages}{38--50}.
\newblock


\bibitem[\protect\citeauthoryear{Gillis}{Gillis}{2014}]%
        {gillis2014and}
\bibfield{author}{\bibinfo{person}{Nicolas Gillis}.}
  \bibinfo{year}{2014}\natexlab{}.
\newblock \showarticletitle{The why and how of nonnegative matrix
  factorization}.
\newblock \bibinfo{journal}{\emph{Regularization, Optimization, Kernels, and
  Support Vector Machines}} \bibinfo{volume}{12}, \bibinfo{number}{257}
  (\bibinfo{year}{2014}).
\newblock


\bibitem[\protect\citeauthoryear{Gonzalez and Zhang}{Gonzalez and
  Zhang}{2005}]%
        {gonzalez2005accelerating}
\bibfield{author}{\bibinfo{person}{Edward~F Gonzalez} {and}
  \bibinfo{person}{Yin Zhang}.} \bibinfo{year}{2005}\natexlab{}.
\newblock \showarticletitle{Accelerating the Lee-Seung algorithm for
  non-negative matrix factorization}.
\newblock \bibinfo{journal}{\emph{Dept. Comput. \& Appl. Math., Rice Univ.,
  Houston, TX, Tech. Rep. TR-05-02}} (\bibinfo{year}{2005}),
  \bibinfo{pages}{1--13}.
\newblock


\bibitem[\protect\citeauthoryear{Hernando, Bobadilla, and Ortega}{Hernando
  et~al\mbox{.}}{2016}]%
        {hernando2016non}
\bibfield{author}{\bibinfo{person}{Antonio Hernando},
  \bibinfo{person}{Jes{\'u}s Bobadilla}, {and} \bibinfo{person}{Fernando
  Ortega}.} \bibinfo{year}{2016}\natexlab{}.
\newblock \showarticletitle{A non negative matrix factorization for
  collaborative filtering recommender systems based on a Bayesian probabilistic
  model}.
\newblock \bibinfo{journal}{\emph{Knowledge-Based Systems}}
  \bibinfo{volume}{97} (\bibinfo{year}{2016}), \bibinfo{pages}{188--202}.
\newblock


\bibitem[\protect\citeauthoryear{Kannan, Ballard, and Park}{Kannan
  et~al\mbox{.}}{2016}]%
        {kannan2016high}
\bibfield{author}{\bibinfo{person}{Ramakrishnan Kannan}, \bibinfo{person}{Grey
  Ballard}, {and} \bibinfo{person}{Haesun Park}.}
  \bibinfo{year}{2016}\natexlab{}.
\newblock \showarticletitle{A high-performance parallel algorithm for
  nonnegative matrix factorization}. In \bibinfo{booktitle}{\emph{ACM SIGPLAN
  Notices}}, Vol.~\bibinfo{volume}{51}. ACM, \bibinfo{pages}{9}.
\newblock


\bibitem[\protect\citeauthoryear{Kim and Park}{Kim and Park}{2008}]%
        {kim2008nonnegative}
\bibfield{author}{\bibinfo{person}{Hyunsoo Kim} {and} \bibinfo{person}{Haesun
  Park}.} \bibinfo{year}{2008}\natexlab{}.
\newblock \showarticletitle{Nonnegative matrix factorization based on
  alternating nonnegativity constrained least squares and active set method}.
\newblock \bibinfo{journal}{\emph{SIAM journal on matrix analysis and
  applications}} \bibinfo{volume}{30}, \bibinfo{number}{2}
  (\bibinfo{year}{2008}), \bibinfo{pages}{713--730}.
\newblock


\bibitem[\protect\citeauthoryear{Kim and Park}{Kim and Park}{2011}]%
        {kim2011fast}
\bibfield{author}{\bibinfo{person}{Jingu Kim} {and} \bibinfo{person}{Haesun
  Park}.} \bibinfo{year}{2011}\natexlab{}.
\newblock \showarticletitle{Fast nonnegative matrix factorization: An
  active-set-like method and comparisons}.
\newblock \bibinfo{journal}{\emph{SIAM Journal on Scientific Computing}}
  \bibinfo{volume}{33}, \bibinfo{number}{6} (\bibinfo{year}{2011}),
  \bibinfo{pages}{3261--3281}.
\newblock


\bibitem[\protect\citeauthoryear{Koitka and Friedrich}{Koitka and
  Friedrich}{2016}]%
        {koitka2016nmfgpu4r}
\bibfield{author}{\bibinfo{person}{Sven Koitka} {and}
  \bibinfo{person}{Christoph~M Friedrich}.} \bibinfo{year}{2016}\natexlab{}.
\newblock \showarticletitle{nmfgpu4R: GPU-Accelerated Computation of the
  Non-Negative Matrix Factorization (NMF) Using CUDA Capable Hardware}.
\newblock \bibinfo{journal}{\emph{R JOURNAL}} \bibinfo{volume}{8},
  \bibinfo{number}{2} (\bibinfo{year}{2016}), \bibinfo{pages}{382--392}.
\newblock


\bibitem[\protect\citeauthoryear{Kuang, Choo, and Park}{Kuang
  et~al\mbox{.}}{2015}]%
        {kuang2015nonnegative}
\bibfield{author}{\bibinfo{person}{Da Kuang}, \bibinfo{person}{Jaegul Choo},
  {and} \bibinfo{person}{Haesun Park}.} \bibinfo{year}{2015}\natexlab{}.
\newblock \showarticletitle{Nonnegative matrix factorization for interactive
  topic modeling and document clustering}.
\newblock In \bibinfo{booktitle}{\emph{Partitional Clustering Algorithms}}.
  \bibinfo{publisher}{Springer}, \bibinfo{pages}{215--243}.
\newblock


\bibitem[\protect\citeauthoryear{Lee and Seung}{Lee and Seung}{2001}]%
        {lee2001algorithms}
\bibfield{author}{\bibinfo{person}{Daniel~D Lee} {and}
  \bibinfo{person}{H~Sebastian Seung}.} \bibinfo{year}{2001}\natexlab{}.
\newblock \showarticletitle{Algorithms for non-negative matrix factorization}.
  In \bibinfo{booktitle}{\emph{Advances in neural information processing
  systems}}. \bibinfo{pages}{556--562}.
\newblock


\bibitem[\protect\citeauthoryear{Liao, Zhang, Guan, and Zhou}{Liao
  et~al\mbox{.}}{2014}]%
        {liao2014cloudnmf}
\bibfield{author}{\bibinfo{person}{Ruiqi Liao}, \bibinfo{person}{Yifan Zhang},
  \bibinfo{person}{Jihong Guan}, {and} \bibinfo{person}{Shuigeng Zhou}.}
  \bibinfo{year}{2014}\natexlab{}.
\newblock \showarticletitle{CloudNMF: a MapReduce implementation of nonnegative
  matrix factorization for large-scale biological datasets}.
\newblock \bibinfo{journal}{\emph{Genomics, proteomics \& bioinformatics}}
  \bibinfo{volume}{12}, \bibinfo{number}{1} (\bibinfo{year}{2014}),
  \bibinfo{pages}{48--51}.
\newblock


\bibitem[\protect\citeauthoryear{Lin}{Lin}{2007}]%
        {lin2007projected}
\bibfield{author}{\bibinfo{person}{Chih-Jen Lin}.}
  \bibinfo{year}{2007}\natexlab{}.
\newblock \showarticletitle{Projected gradient methods for nonnegative matrix
  factorization}.
\newblock \bibinfo{journal}{\emph{Neural computation}} \bibinfo{volume}{19},
  \bibinfo{number}{10} (\bibinfo{year}{2007}), \bibinfo{pages}{2756--2779}.
\newblock


\bibitem[\protect\citeauthoryear{Liu, Yang, Fan, He, and Wang}{Liu
  et~al\mbox{.}}{2010}]%
        {liu2010distributed}
\bibfield{author}{\bibinfo{person}{Chao Liu}, \bibinfo{person}{Hung-chih Yang},
  \bibinfo{person}{Jinliang Fan}, \bibinfo{person}{Li-Wei He}, {and}
  \bibinfo{person}{Yi-Min Wang}.} \bibinfo{year}{2010}\natexlab{}.
\newblock \showarticletitle{Distributed nonnegative matrix factorization for
  web-scale dyadic data analysis on mapreduce}. In
  \bibinfo{booktitle}{\emph{Proceedings of the 19th international conference on
  World wide web}}. ACM, \bibinfo{pages}{681--690}.
\newblock


\bibitem[\protect\citeauthoryear{Lopes and Ribeiro}{Lopes and Ribeiro}{2010}]%
        {lopes2010non}
\bibfield{author}{\bibinfo{person}{Noel Lopes} {and}
  \bibinfo{person}{Bernardete Ribeiro}.} \bibinfo{year}{2010}\natexlab{}.
\newblock \showarticletitle{Non-negative matrix factorization implementation
  using graphic processing units}. In \bibinfo{booktitle}{\emph{International
  Conference on Intelligent Data Engineering and Automated Learning}}.
  Springer, \bibinfo{pages}{275--283}.
\newblock


\bibitem[\protect\citeauthoryear{Mej{\'\i}a-Roa, Tabas-Madrid, Setoain,
  Garc{\'\i}a, Tirado, and Pascual-Montano}{Mej{\'\i}a-Roa
  et~al\mbox{.}}{2015}]%
        {mejia2015nmf}
\bibfield{author}{\bibinfo{person}{Edgardo Mej{\'\i}a-Roa},
  \bibinfo{person}{Daniel Tabas-Madrid}, \bibinfo{person}{Javier Setoain},
  \bibinfo{person}{Carlos Garc{\'\i}a}, \bibinfo{person}{Francisco Tirado},
  {and} \bibinfo{person}{Alberto Pascual-Montano}.}
  \bibinfo{year}{2015}\natexlab{}.
\newblock \showarticletitle{NMF-mGPU: non-negative matrix factorization on
  multi-GPU systems}.
\newblock \bibinfo{journal}{\emph{BMC bioinformatics}} \bibinfo{volume}{16},
  \bibinfo{number}{1} (\bibinfo{year}{2015}), \bibinfo{pages}{43}.
\newblock


\bibitem[\protect\citeauthoryear{Robila and Maciak}{Robila and Maciak}{2006}]%
        {robila2006parallel}
\bibfield{author}{\bibinfo{person}{Stefan~A Robila} {and}
  \bibinfo{person}{Lukasz~G Maciak}.} \bibinfo{year}{2006}\natexlab{}.
\newblock \showarticletitle{A parallel unmixing algorithm for hyperspectral
  images}. In \bibinfo{booktitle}{\emph{Intelligent Robots and Computer Vision
  XXIV: Algorithms, Techniques, and Active Vision}},
  Vol.~\bibinfo{volume}{6384}. International Society for Optics and Photonics,
  \bibinfo{pages}{63840F}.
\newblock


\bibitem[\protect\citeauthoryear{Shi, Kang, Choo, and Reddy}{Shi
  et~al\mbox{.}}{2018}]%
        {shi2018short}
\bibfield{author}{\bibinfo{person}{Tian Shi}, \bibinfo{person}{Kyeongpil Kang},
  \bibinfo{person}{Jaegul Choo}, {and} \bibinfo{person}{Chandan~K Reddy}.}
  \bibinfo{year}{2018}\natexlab{}.
\newblock \showarticletitle{Short-Text Topic Modeling via Non-negative Matrix
  Factorization Enriched with Local Word-Context Correlations}. In
  \bibinfo{booktitle}{\emph{Proceedings of the 2018 World Wide Web Conference
  on World Wide Web}}. International World Wide Web Conferences Steering
  Committee, \bibinfo{pages}{1105--1114}.
\newblock


\bibitem[\protect\citeauthoryear{Smith et~al\mbox{.}}{Smith
  et~al\mbox{.}}{2018}]%
        {smith2018theory}
\bibfield{author}{\bibinfo{person}{Tyler~Michael Smith} {et~al\mbox{.}}}
  \bibinfo{year}{2018}\natexlab{}.
\newblock \emph{\bibinfo{title}{Theory and practice of classical matrix-matrix
  multiplication for hierarchical memory architectures}}.
\newblock \bibinfo{thesistype}{Ph.D. Dissertation}.
\newblock


\bibitem[\protect\citeauthoryear{Suh, Choo, Lee, and Reddy}{Suh
  et~al\mbox{.}}{2017}]%
        {suh2017local}
\bibfield{author}{\bibinfo{person}{Sangho Suh}, \bibinfo{person}{Jaegul Choo},
  \bibinfo{person}{Joonseok Lee}, {and} \bibinfo{person}{Chandan~K Reddy}.}
  \bibinfo{year}{2017}\natexlab{}.
\newblock \showarticletitle{Local topic discovery via boosted ensemble of
  nonnegative matrix factorization}. In \bibinfo{booktitle}{\emph{Proceedings
  of the 26th International Joint Conference on Artificial Intelligence}}. AAAI
  Press, \bibinfo{pages}{4944--4948}.
\newblock


\bibitem[\protect\citeauthoryear{Wang, Wang, and Gao}{Wang
  et~al\mbox{.}}{2013}]%
        {wang2013non}
\bibfield{author}{\bibinfo{person}{Jim Jing-Yan Wang}, \bibinfo{person}{Xiaolei
  Wang}, {and} \bibinfo{person}{Xin Gao}.} \bibinfo{year}{2013}\natexlab{}.
\newblock \showarticletitle{Non-negative matrix factorization by maximizing
  correntropy for cancer clustering}.
\newblock \bibinfo{journal}{\emph{BMC bioinformatics}} \bibinfo{volume}{14},
  \bibinfo{number}{1} (\bibinfo{year}{2013}), \bibinfo{pages}{107}.
\newblock


\bibitem[\protect\citeauthoryear{Yang and Michailidis}{Yang and
  Michailidis}{2015}]%
        {yang2015non}
\bibfield{author}{\bibinfo{person}{Zi Yang} {and} \bibinfo{person}{George
  Michailidis}.} \bibinfo{year}{2015}\natexlab{}.
\newblock \showarticletitle{A non-negative matrix factorization method for
  detecting modules in heterogeneous omics multi-modal data}.
\newblock \bibinfo{journal}{\emph{Bioinformatics}} \bibinfo{volume}{32},
  \bibinfo{number}{1} (\bibinfo{year}{2015}), \bibinfo{pages}{1--8}.
\newblock


\bibitem[\protect\citeauthoryear{Zhang, Wang, Ford, and Makedon}{Zhang
  et~al\mbox{.}}{2006}]%
        {zhang2006learning}
\bibfield{author}{\bibinfo{person}{Sheng Zhang}, \bibinfo{person}{Weihong
  Wang}, \bibinfo{person}{James Ford}, {and} \bibinfo{person}{Fillia Makedon}.}
  \bibinfo{year}{2006}\natexlab{}.
\newblock \showarticletitle{Learning from incomplete ratings using non-negative
  matrix factorization}. In \bibinfo{booktitle}{\emph{Proceedings of the 2006
  SIAM international conference on data mining}}. SIAM,
  \bibinfo{pages}{549--553}.
\newblock


\end{thebibliography}

\end{document}